\documentclass[10pt,twocolumn,letterpaper]{article}

\usepackage[pagenumbers]{cvpr}

\usepackage{pifont}

\usepackage{siunitx}
\usepackage{makecell}
\usepackage{graphicx}
\usepackage{multirow}
\usepackage{setspace}
\usepackage{dcolumn}
\usepackage{comment}
\usepackage{colortbl}
\usepackage{pifont}
\usepackage{placeins}

\usepackage[pagebackref,breaklinks,colorlinks]{hyperref}
\definecolor{darkgreen}{rgb}{0.0, 0.5, 0.0}

\def\method{Steer3D\xspace}
\def\bench{\textsc{Edit3D-Bench}\xspace}
\def\trellis{TRELLIS\xspace}
\def\edittrellis{Edit-TRELLIS\xspace}

\newcommand{\mypar}[1]{\vspace{1mm}\noindent\textbf{#1}}
\newcommand{\myparit}[1]{\vspace{1mm}\noindent\emph{#1}}

\title{Feedforward 3D Editing via Text-Steerable Image-to-3D}

\author{Ziqi Ma
\qquad
Hongqiao Chen
\qquad
Yisong Yue
\qquad
Georgia Gkioxari
\\
California Institute of Technology
}

\begin{document}

\twocolumn[{%
\renewcommand\twocolumn[1][]{#1}%
\maketitle
\begin{center}
    \vspace{-5mm}
    \centering
    \includegraphics[width=\linewidth]{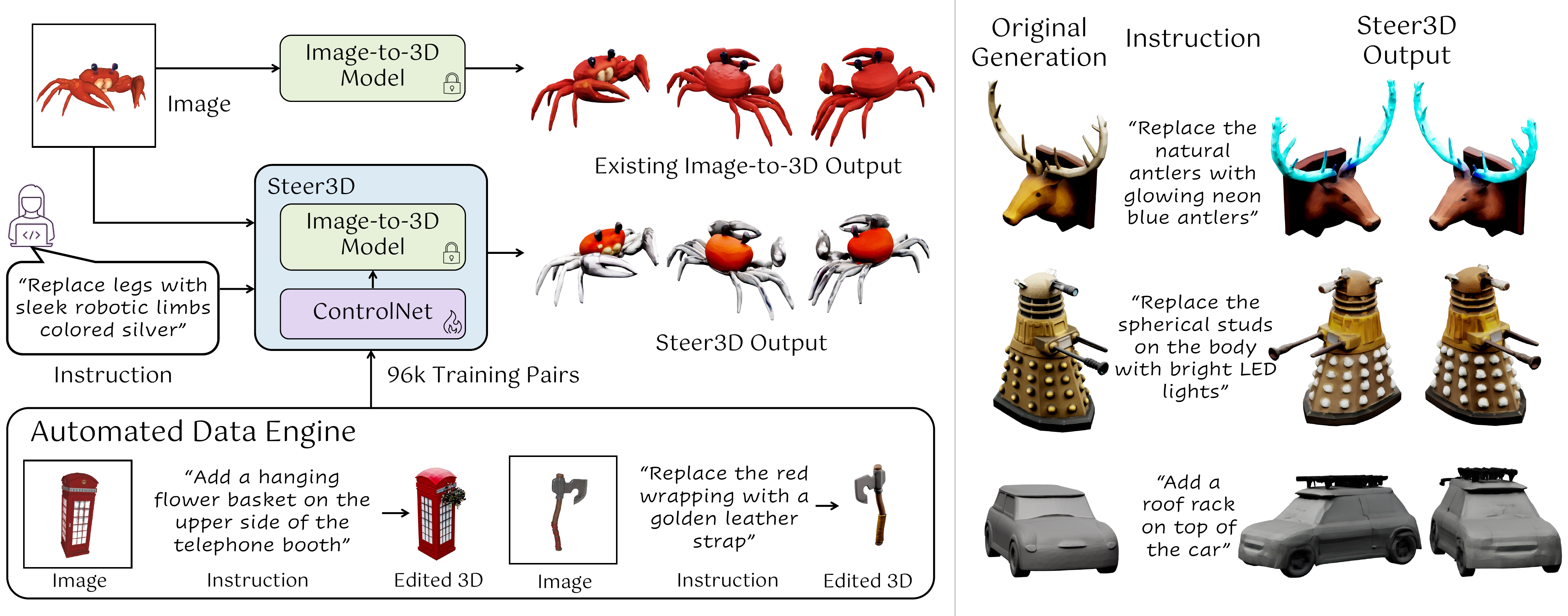}
    \vspace{-5mm}
    \captionof{figure}{We present \method, which enables feedforward editing of generated 3D assets by ingesting text steerability to image-to-3D models. \method uses an architecture inspired by ControlNet to leverage image-to-3D pretraining and achieve data efficiency. We build an automated data engine that generates $96k$ synthetic editing pairs for scalable training. \method shows strong editing capabilities while staying consistent with the original 3D asset, as shown on the right side.
    }
    \label{fig:teaser}
\end{center}
}]

\begin{abstract}
Recent progress in image-to-3D has opened up immense possibilities for design, AR/VR, and robotics. However, to use AI-generated 3D assets in real applications, a critical requirement is the capability to edit them easily.
We present a feedforward method, \method, to add text steerability to image-to-3D models, which enables editing of generated 3D assets with language. Our approach is inspired by ControlNet, which we adapt to image-to-3D generation to enable text steering directly in a forward pass.
We build a scalable data engine for automatic data generation, and develop a two-stage training recipe based on flow-matching training and Direct Preference Optimization (DPO).
 Compared to competing methods, \method more faithfully follows the language instruction and maintains better consistency with the original 3D asset, while being $2.4\times$ to $28.5\times$ faster. \method demonstrates that it is possible to add a new modality (text) to steer the generation of pretrained image-to-3D generative models with $100k$ data. Project website: \url{https://glab-caltech.github.io/steer3d/}
\end{abstract}    
\section{Introduction}
\label{sec:intro}

Generative image-to-3D models~\citep{xiang_structured_2024, zhao_hunyuan3d_2025, hong_lrm_2024} enable users to create 3D assets from a single image, unlocking new applications in design, AR/VR, and robotic simulation. However, integrating AI generations into real applications requires editing and customizing the generated 3D assets, a capability not supported by these models. 

Existing solutions to 3D editing include 2D-3D pipelines which use image editing to change the object views and then lift to 3D \citep{chen_dge_2024, qi_tailor3d_2024, wang_view-consistent_2025}. However, they suffer from inconsistent image edit outputs, which are reflected in the reconstructed 3D shape. These methods are also slow -- they take several minutes per edit.
To address these inherent limitations with pipeline methods, we want to train a feedforward model that directly perform editing in 3D.
The most intuitive approach is to design a generative model which takes the original 3D asset as input, and generates the edited asset conditioned on the edit instruction. However, such models rely on paired 3D editing data with text instructions, which is very difficult to obtain. Due to this challenge, training such a model from scratch is impractical.

We propose a solution from a different perspective: instead of training an editing model from scratch, we can turn any image-to-3D model into an editing model by augmenting it with text steerability. \cref{fig:teaser} shows an overview of our method, \method. We start from a pretrained image-to-3D generator and augment it with ControlNet \citep{zhang_adding_2023}. ControlNet ingests the steering prompt and guides the base model toward the desired 3D edit. To edit any asset generated by the base model, we can run \method with the input image plus the editing text, and directly obtain an edited asset that follows the instruction and remains consistent with the original asset. By leveraging shape and object priors from pretrained image-to-3D models, this design is more data-efficient than training a standalone editing model from scratch. 

A data-efficient architecture makes feedforward 3D editing tractable, yet we still need to collect paired 3D edit data for training. We develop an automated data engine that combines image editing models, large vision–language models, and image-to-3D generators to synthesize diverse geometry- and texture-edit pairs. In total, we produce $96$k training pairs covering a wide range of shape and appearance changes. We then train \method with flow matching \citep{lipman_flow_2023}, and apply Direct Preference Optimization (DPO; \cite{rafailov_direct_2023}) to avoid the trivial ``no-edit'' solution.
Our analysis shows that both the data engine and the training recipe scale effectively.
\cref{fig:teaser} (right) shows outputs from \method.

Our main contributions are as follows:
\begin{itemize}
\item We build a feedforward model for 3D editing which demonstrates stronger instruction following and better consistency preservation than existing methods, while being $2.4\times$ to $28.5\times$ faster.
\item We propose a novel architecture that adapts ControlNet to image-to-3D generation to enable language steering.
\item We design a scalable, automatic data engine that can generate 3D editing paired data.
\item We develop a two-stage training recipe that integrates flow-matching and Direct Preference Optimization to add text steering capability to pretrained image-to-3D models.
\item We release \bench, a 3D editing benchmark with diverse object types and editing instructions.
\end{itemize}

\begin{figure*}[t]
\begin{center}
\centerline{\includegraphics[width=2\columnwidth]{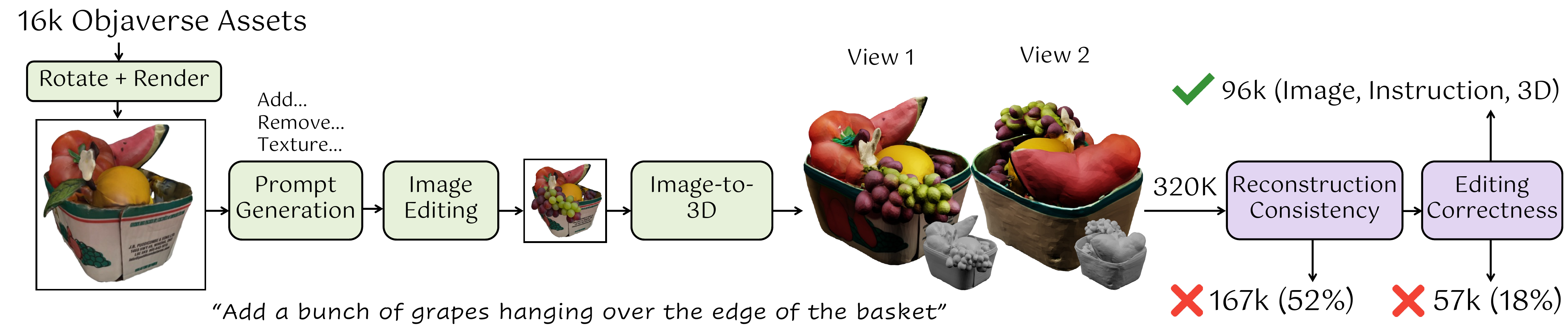}}
\caption{\method data engine: we begin with one rendered view for each Objaverese object and query an LLM for diverse editing instructions. Then, a 2D editing model is used to generate an edited view, which is subsequently reconstructed into a 3D asset. Editing pairs pass through a two-stage filter, in which approximately 30\% of all generated pairs are kept.}
\label{fig:dataengine}
\end{center}
\vspace{-6mm}
\end{figure*}
\section{Related Work}
\label{sec:relatedworks}

\mypar{Image-to-3D Models.}
Image-to-3D models generate 3D assets based on a single image. Recent state-of-art models \citep{xiang_structured_2024, zhao_hunyuan3d_2025, wu_unilat3d_2025} formulate image-to-3D as conditional generation via rectified flow~\cite{lipman_flow_2023}. These models don't allow direct text-based editing, even though \trellis~\citep{xiang_structured_2024} can be combined with inpainting methods \cite{lugmayr2023inpainting} to perform limited editing, assuming the additional input of a 3D bounding box. We study editing with language guidance and no additional input, and via a forward pass.

\mypar{3D Editing Pipelines Based on 2D Editing.}
Starting from Instruct-NeRF2NeRF \citep{haque_instruct-nerf2nerf_2023}, various methods \citep{chen_generic_2024, chen_dge_2024, erkoc_preditor3d_2024, wu_gaussctrl_2024, khalid_latenteditor_2024, chen_consistdreamer_2024} build pipelines with 2D editing models to perform 3D editing. \citep{haque_instruct-nerf2nerf_2023, hong_perturb-and-revise_2025} are based on NeRF~\citep{mildenhall2021nerf}, \citep{haque_instruct-nerf2nerf_2023, hong_perturb-and-revise_2025, wang_view-consistent_2025} are based on Gaussian splatting~\citep{kerbl20233d}, and \citep{qi_tailor3d_2024} is based on tri-plane. While there is active effort on improving multiview editing consistency ~\citep{chen_generic_2024, karim_free-editor_2024, wang_view-consistent_2025}, this issue is still challenging, limiting editing quality. Furthermore, these pipelines are also slow due to rendering, multiple model invocations and NeRF or Gaussian optimization. 

\mypar{Test-Time Optimization.}
Test-time optimization for 3D editing includes score-distillation-based methods and inversion-based methods. Score distillation \citep{poole_dreamfusion_2022} is adapted for 3D editing in ~\citep{sella_vox-e_2023, zhuang_tip-editor_2024, he_customize_2023, Kim_2025_ICCV}. These methods use the score distillation loss from 2D text-to-image models to guide optimization. 3D-LATTE \citep{parelli_3d-latte_2025} is an inversion-based method that uses attention injection. These methods are slow due to the optimization required, and oftentimes contain instance-specific hyperparameters.

\mypar{Feedforward 3D Editing.}
Feedforward 3D editing is very challenging due to the lack of large-scale 3D edit pairs. SHAP-EDITOR \citep{chen2024shap} trains \textit{one model per editing instruction}, and only supports 6 editing instructions. Other methods tackle an easier version of 3D editing, assuming the extra input of 3D bounding box, such as MaskedLRM~\citep{gao_3d_2025} and Instant3DEdit~\cite{barda2025instant3dit}.
Another class of feedforward 3D editing models is LLM-based, such as ShapeLLM-Omni \citep{ye_shapellm-omni_2025} which tokenizes 3D assets, or LL3M~\citep{lu_ll3m_2025} and BlenderAlchemy~\citep{huang_blenderalchemy_2024}, which are agentic frameworks that edits blender code of 3D objects.
\section{Method}

We enable editing of 3D assets by augmenting image-to-3D models with text steerability. Given a reference image of an object and a text instruction, we design a feedforward model that generates a new asset which follows the text instruction while being faithful to the ``unsteered" generation of the image-to-3D model. To train such a model, we introduce three key components: (a) an architecture design that can leverage generative pretraining of image-to-3D models; (b) an automatic data engine that generates creative editing prompts and pre-post editing pairs; (c) a two-stage training recipe combining flow-matching training and Direct Preference Optimization (DPO; \citep{rafailov_direct_2023}).

\subsection{Architecture}
\begin{figure}[t]
\vspace{-3mm}
\begin{center}
\centerline{\includegraphics[width=\columnwidth]{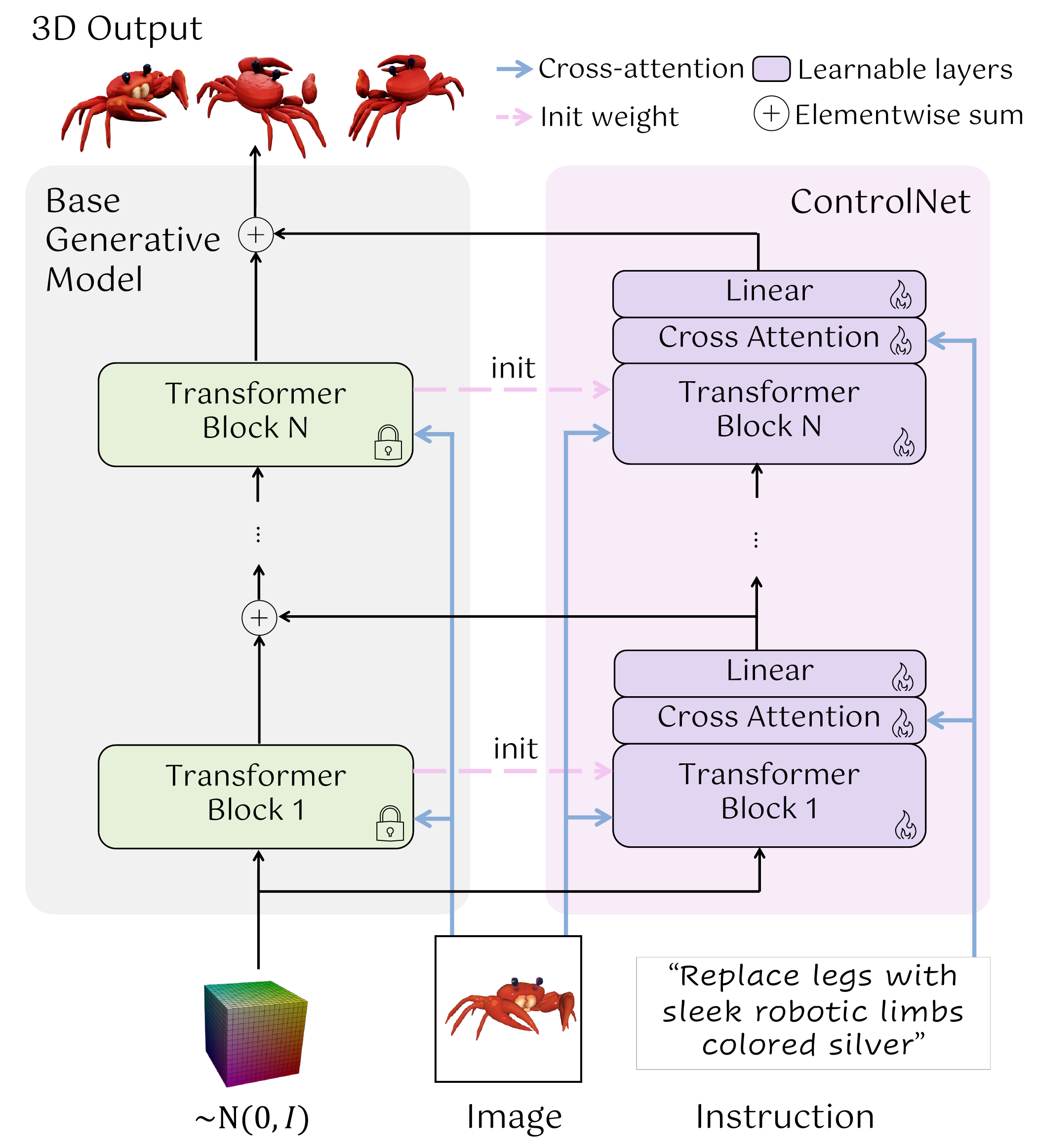}}
\vspace{-2mm}
\caption{\method architecture: we design a ControlNet-based architecture to leverage the shape and geometry prior of pretrained image-to-3D generative models. We add a trainable ControlNet block corresponding to each transformer block in the base model.}
\label{fig:controlnet}
\vspace{-13mm}
\end{center}
\end{figure}

Existing image-to-3D generative methods \citep{xiang_structured_2024,zhao_hunyuan3d_2025,yang_hunyuan3d_2025} rely on large-scale pretraining. For example, \trellis~\citep{xiang_structured_2024} is trained on $500k$ (image, 3D) pairs. To give them an additional 3D editing capability with natural language instructions is expected to require even more training data, in the form of (image, instruction, 3D) triplets. Such data is hard to obtain at scale.
We propose an alternative that is less data-hungry, inspired by advances in image editing \citep{zhang_adding_2023}. We build upon existing pretrained 3D models, and learn to steer their representations by designing a text-conditioned 3D ControlNet, as shown in \cref{fig:controlnet}. 

\mypar{Base Model.}
\method uses a ControlNet architecture \citep{zhang_adding_2023} adapted to image-to-3D generation. We choose \trellis \citep{xiang_structured_2024}, a state-of-the-art image-to-3D approach, as our base model. \trellis is composed of two rectified flow models: the first flow model generates coarse geometry represented as binary occupancy in a $64 \times 64 \times 64$ voxel grid. The second flow model takes the generated geometry as input, and generates latent features that can decode into Gaussian splats, radiance field, or a mesh. We apply the same architecture to both the geometry and the texture model of \trellis.

\mypar{ControlNet for 3D Editing.}
The base model is composed of 24 stacked transformer blocks. We add a ControlNet block corresponding to each layer: we first copy the architecture and weights of the base model transformer block, then add additional cross attention that attends to the editing text, and append a projection layer that is initialized at zero. The output of each ControlNet block is added to the output of the corresponding base model block, which is then passed to the next transformer block of the base model. The ControlNet output is also passed to the next ControlNet block. This is illustrated in~\cref{fig:controlnet}.

The base model is kept frozen, and only ControlNet components are trainable. Since each projection layer is zero-initialized, the initial outputs of all ControlNet blocks are zero, meaning \method predicts the base model output at initialization. This provides us with a good optimization starting point. Furthermore, since the ControlNet transformer blocks are initialized with the base model weights, the control branch preserves object and shape priors learned during the base model's pretraining. These design choices enable data-efficient, stable training.
We deviate from the the original ControlNet paper \citep{zhang_adding_2023}. ControlNet was originally UNet-based and added control to skip connections. We adapt it for transformer-based flow models and add it directly to each block's feature.

\subsection{Data Engine}
Feedforward models are data-driven. Yet, relying on humans to generate large-scale 3D editing pair data is infeasible. To tackle this problem, we propose a scalable synthetic data pipeline utilizing 2D editing models and image-to-3D generative models. We sample $16k$ objects from Objaverse \citep{deitke_objaverse-xl_2023} and generate $96k$ high quality 3D editing pairs.

Our data engine is illustrated in \cref{fig:dataengine}. For each Objaverse asset, we first rotate the object and render a 2D view, so that our data is not constrained to front-side edits. Next, we prompt a VLM, GPT-4.1-mini, to propose a total of 20 creative edits in the following categories: addition, removal, and new texture. Then, we edit the rendered view with a state-of-the-art 2D editing model, Step1X-Edit \citep{liu_step1x-edit_2025}, and reconstruct to 3D using an image-to-3D model, Hunyuan-3D 2.1 \citep{zhao_hunyuan3d_2025}. Our data engine generates $320k$ raw 3D pairs using a total of $2,500$ H100 hours.

To improve data quality, we perform two-stage data filtering to reduce incorrect edits and inconsistent reconstructions. For image editing errors, we first prompt a VLM, unaware of the editing instruction, to list the differences between a rendered view of the original and edited asset. Then, we query a separate LLM (without giving it the images) to determine if the differences listed indicate a successful and accurate edit. We separate these two roles to reduce hallucination.
For reconstruction inconsistency, we use 2D perceptual similarity \citep{fu_dreamsim_2023} thresholding on rendered images of the generated asset and the edited image, and discard pairs with low similarity. 
This two-stage filter filters out $70\%$ of all generated pairs. We show outputs of our data engine in \cref{fig:teaser} and \cref{fig:benchmark}.

\subsection{Training Recipe}
\method uses a two-stage training recipe: (1) supervised flow-matching training; (2) Direct Preference Optimization (DPO; \citep{rafailov_direct_2023}) to penalize the ``no edit'' behavior, a local optimum for our model.

\mypar{Supervised Flow-Matching Training.}
Similar to our base model, \method is a rectified flow model which defines a linear forward process between data samples $x_0$ and noise samples $\epsilon \sim \mathcal{N}(0{,} I)$, $x(t) = (1-t)x_0 +t\epsilon$. The flow-matching objective aims to align the velocity $v$ at each timestep with the overall direction of $\epsilon - x_0$.
We start by fine-tuning the ControlNet weights with the text prompt via the normal flow-matching loss:
\vspace{-1mm}
\begin{align*}
L_{\phi}^\text{SFT}=\mathbb{E}_{t,x_0,\epsilon}[\Vert v_{\theta, \phi}(x_t,t) - (\epsilon - x_0) \Vert^2]
\end{align*}
\vspace{-1mm}
 where $\theta$ is the frozen base model weights and $\phi$ is the trainable ControlNet weights, $v_{\theta, \phi}$ is the predicted velocity by the ControlNet-enhanced model.

\mypar{Direct Preference Optimization.}
Supervised flow-matching training can yield a conservative model which sometimes ignores the edit prompt and predicts the unedited asset. We hypothesize that this is because the pre- and post-edit assets are close in the latent space for localized edits, and a distribution-matching-type objective might struggle to push the prediction apart from the original output (which is also our initialization based on ControlNet). We want to explicitly discourage the ``no edit" behavior via DPO. For each editing pair, we use the ground truth as a positive example, the original generation as a negative example, and apply the DPO loss adapted for flow matching \citep{liu2025improving}:

\vspace{-8mm}
{\small
\begin{align*}
& L_{\phi}^\text{DPO} = -\mathbb{E}[\log \text{sigmoid}
(-\frac{\beta}{2} \\
& ( (\Vert v_{\theta, \phi}(x_+,t) - (\epsilon - x_+) \Vert^2 -
\Vert v_{\theta, \phi}(x_-,t) - (\epsilon - x_-) \Vert^2) - \\
& (\Vert v_\text{ref}(x_+,t) - (\epsilon - x_+) \Vert^2 - \Vert v_\text{ref}(x_-,t) - (\epsilon - x_-) \Vert^2))]\\
& + \alpha L_{\phi}^\text{SFT}
\end{align*}
}
where $x_+$ is the edited (positive) asset latent, and $x_-$ the unedited (negative) asset latent, and $v_\text{ref}$ is the checkpoint from the supervised flow-matching training. To stabilize DPO training, we also add the flow-matching loss as regularization.

\mypar{Training Details.}
Since our base model, \trellis, contains two separate flow models for geometry and texture respectively, we perform separate training for each stage, following roughly the same recipe.
For both stages, we apply gradient clipping to stabilize training. We sample timesteps with a higher standard deviation than the original \trellis training, to increase coverage at small timesteps.
For the geometry ControlNet, we train separately for addition and removal for the best performance, and do not perform DPO. For texture editing, since geometry should be unchanged, the geometry model is run with no control, yielding the source geometry. We train on 6 A100 GPUs with per-GPU batch size of 2 and gradient accumulation of 2 for the geometry stage, and per-GPU batch size of 8 without gradient accumulation for the texture stage. We note that \method separates geometry and texture because our base model, \trellis, separates them into two models. However, our recipe is not limited to two-stage models.

\mypar{Classifier-Free Guidance.}
Similar to other flow-based models, we apply classifier-free guidance (CFG; \cite{ho_classifier-free_2022}) during supervised flow-matching training for the texture ControlNet by dropping the editing text with a probability of 0.2. For the geometry model, we train without CFG because it can lead to unstable training. At inference time, we can still optionally perform CFG by using the original base model (which does not take in text) as the unconditional model.
\section{\bench}

3D editing has been largely evaluated qualitatively. Since different methods assume different inputs, such as multiple posed views~\cite{haque_instruct-nerf2nerf_2023, chen_generic_2024, wang_view-consistent_2025}, optimized Gaussian splats~\citep{chen_dge_2024}, or NeRF~\citep{hong_perturb-and-revise_2025}, there is no standard benchmark. 
Prior methods evaluate editing quality based on metrics such as CLIPScore \citep{hessel_clipscore_2022} or PickScore \citep{kirstain_pick--pic_2023}, or LLM-based text-to-3D evaluator Eval3D \citep{duggal_eval3d_2025}.
These metrics, not designed for 3D editing evaluation, have 3 key limitations: (1) they are designed for descriptive, caption-type text, and editing prompts such as ``add something", ``remove something", ``replace something" are out of distribution; (2) they only measure alignment to the editing text, but cannot evaluate consistency with the original 3D asset. (3) they rely on 2D renderings and cannot directly evaluate 3D geometry.

\begin{figure}[t]
\begin{center}
\centerline{\includegraphics[width=\columnwidth]{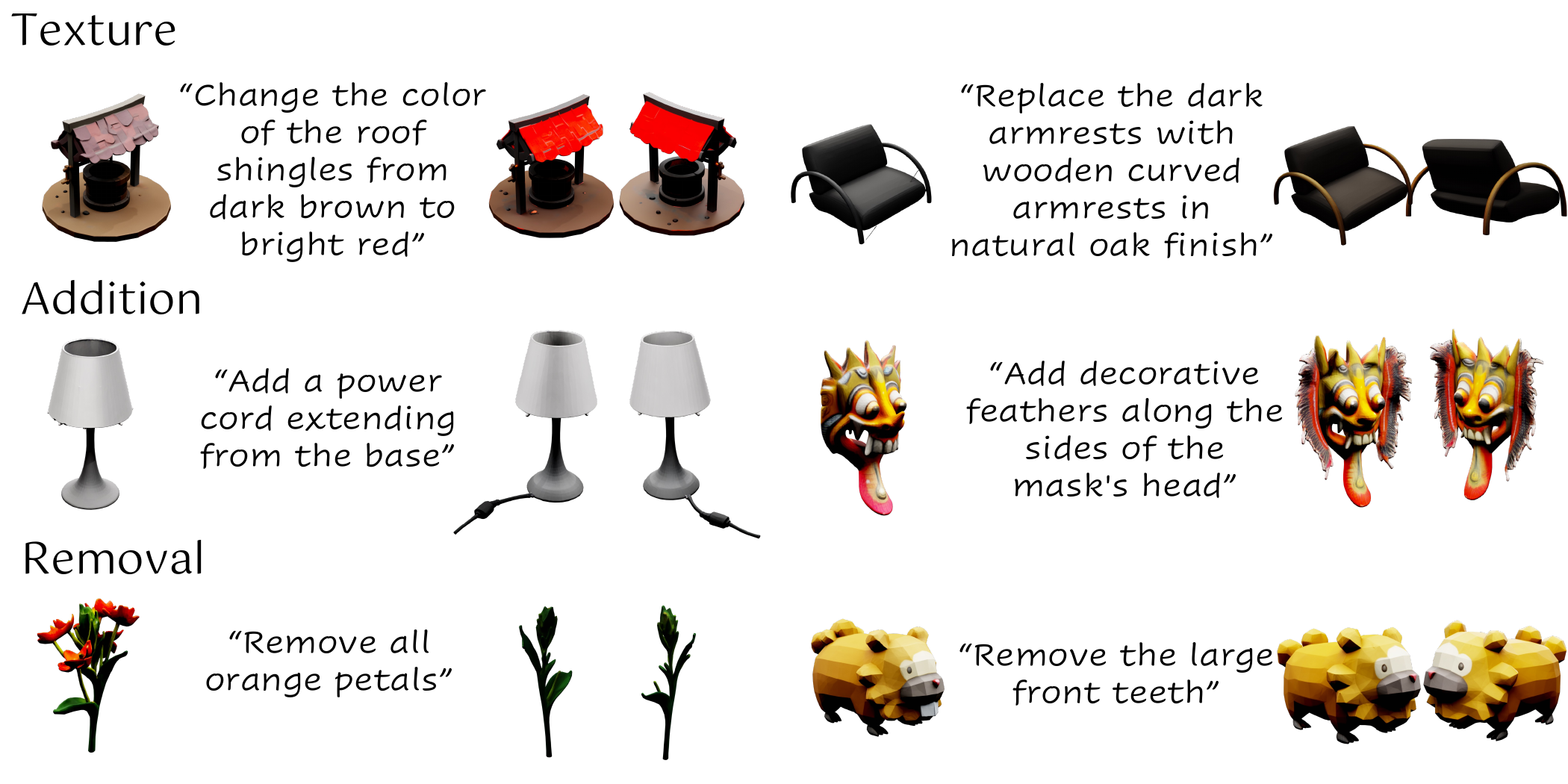}}
\caption{We present \bench, which contains diverse edits spanning texture, addition, and removal. It comprises 250 objects and 250 distinct edits, which is $41.6 \times$ the size of the existing commonly-evaluated objects from InstructNerf2Nerf.}
\label{fig:benchmark}
\end{center}
\vspace{-12mm}
\end{figure}


We propose \bench, a benchmark composed of (pre-edit 3D, instruction, post-edit 3D) triplets. The ground truth post-edit 3D enables evaluation of both correctness and consistency, via 3D geometry metrics such as Chamfer Distance and F1 score, and 2D perceptual metrics such as LPIPS~\citep{zhang2018unreasonable} on rendered views.

\bench is sourced from our data engine and selected and verified by humans. \bench encompasses 250 objects with distinct and detailed edit instructions, including 150 distinct texture edits, 50 addition edits and 50 removal edits. \bench is $41.6\times$ the size of the existing commonly-evaluated examples. Examples are shown in~ \cref{fig:benchmark}. Our instructions are more detailed and diverse than exisitng commonly-evaluated instructions such as ``add a hat" or ``man to clown". We also provide tooling for 3D representation conversion and metric calculation, making the benchmark plug-and-play.
\section{Experiments}

\begin{figure*}[t]
\begin{center}
\vspace{-15pt}
\centerline{\includegraphics[width=\textwidth]{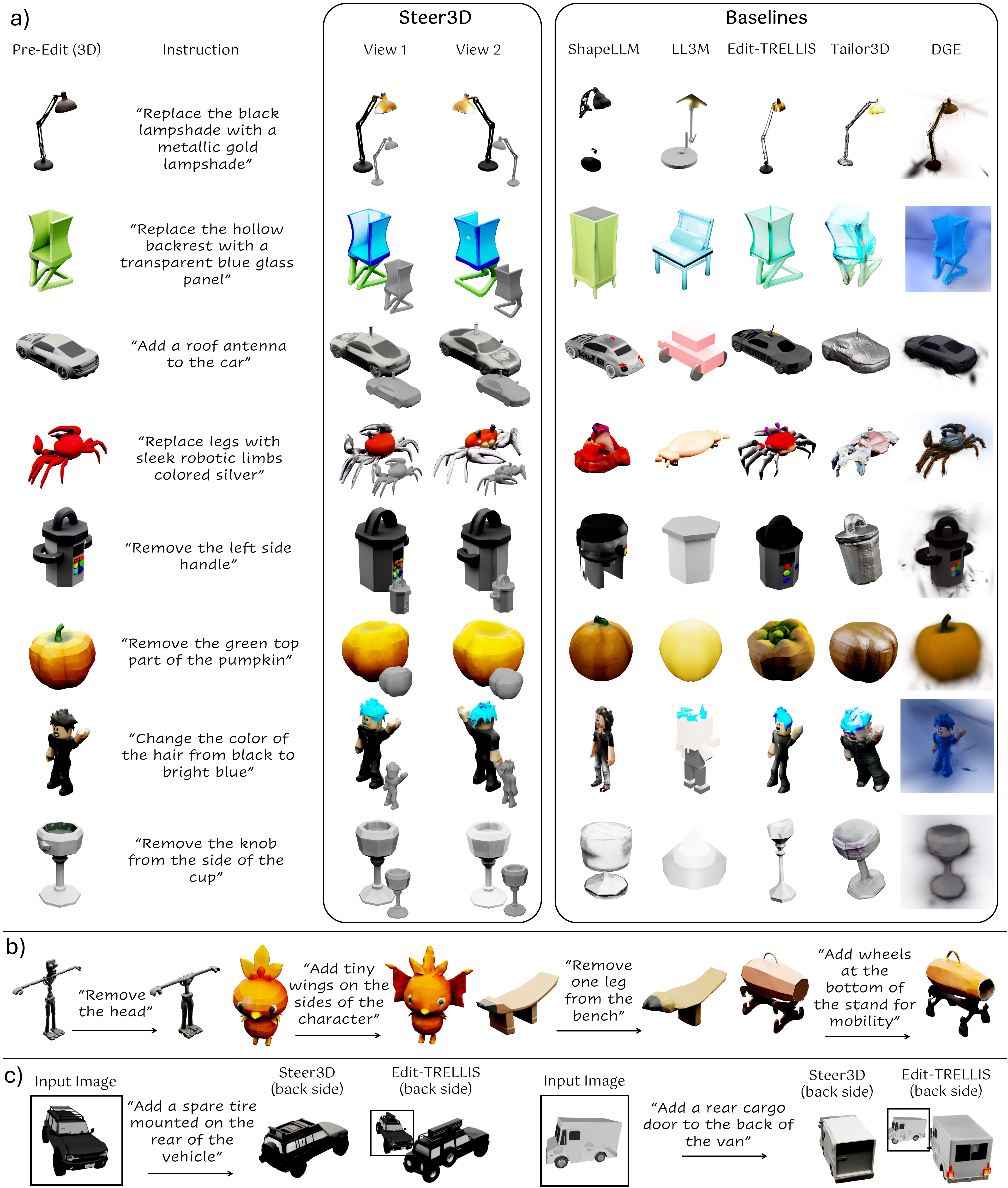}}

\caption{(a): Qualitative comparison of our method with baselines. \method shows strong advantage both in editing correctness and in consistency with the pre-edit 3D asset. \method demonstrates strong localization capability -- e.g., editing only the chair’s hollow backrest, while baselines like \edittrellis and LL3M fail by modifying the whole chair. (b): More qualitative examples of \method. (c): \method is able to perform edits that are not visible in the input view, whereas \edittrellis, which relies on front-view 2D editing, fails.}
\label{fig:qualitative}
\vspace{-15pt}
\end{center}
\end{figure*}


\begin{table*}[h]
\begin{center}
\resizebox{\linewidth}{!}{
\begin{tabular}
{l|ccc|ccc|ccc|ccc}
\toprule
 & \multicolumn{6}{c|}{Removal} & \multicolumn{6}{c}{Addition} \\
 \midrule
 & \multicolumn{3}{c|}{Seen Assets, Unseen Edits} & \multicolumn{3}{c|}{Unseen Assets} & \multicolumn{3}{c|}{Seen Assets, Unseen Edits} & \multicolumn{3}{c}{Unseen Assets} \\
\midrule
  & LPIPS ($\downarrow$)  & Chamfer ($\downarrow$) & F1 ($\uparrow$) & LPIPS  ($\downarrow$) & Chamfer ($\downarrow$) &F1 ($\uparrow$) & LPIPS ($\downarrow$)  & Chamfer ($\downarrow$) & F1 ($\uparrow$) & LPIPS  ($\downarrow$) & Chamfer ($\downarrow$) &F1 ($\uparrow$) \\
\midrule
Tailor3D & 0.234 & 0.154 & 0.117 & 0.218 & 0.196 & 0.138 &  0.259 & 0.160 & 0.116 & 0.265 & 0.152 & 0.126 \\
{\edittrellis} & 0.192 & 0.133 & 0.189 & 0.169 & 0.218 & 0.144 & 0.227 & 0.196 & 0.179 & 0.236 & 0.171 & 0.183 \\
 DGE & 0.219 & 0.235 & 0.093 & 0.168 & 0.262 & 0.080 & 0.229 & 0.245 & 0.063 & 0.227  & 0.234 & 0.077 \\
ShapeLLM & 0.221 &0.147 & 0.181 & 0.203 & 0.210 & 0.128 & 0.265 & 0.162 & 0.158 & 0.265 & 0.180 & 0.152 \\
 {\method} (Ours) & \textbf{0.168} & \textbf{0.049} & \textbf{0.310} & \textbf{0.146} & \textbf{0.102} & \textbf{0.261} & \textbf{0.181} & \textbf{0.107} & \textbf{0.285} & \textbf{0.199} & \textbf{0.089} & \textbf{0.253} \\
\bottomrule
\end{tabular}
} 
\end{center}
\vspace{-4mm}
\caption{Quantitative comparison on geometry edits: addition and removal. We evaluate LPIPS, Chamfer Distance, and F1 score both on seen assets (unseen edits) and unseen assets. \method shows strong advantage on all metrics in all settings, with 64\% higher F1 score, 63\% reduction in Chamfer Distance, and 53\% reduction in LPIPS.}
\label{table:geometry}
\vspace{-5mm}
\end{table*}


We conduct both qualitative and quantitative evaluations to assess the 3D editing capabilities of \method. We show that \method outperforms competing approaches qualitatively and quantitatively on both geometry and appearance-based edits, while being $2.4\times$ to $28.5 \times$ faster. Finally, we include ablation studies and a scaling analysis that validate key design and training choices.

\mypar{Baselines.}
We compare \method with both feedforward methods that directly edit in 3D, and pipeline methods that leverage 2D editing.

\myparit{\textbf{Feedforward baselines:}}
\begin{itemize}
\item ShapeLLM-Omni~\citep{ye_shapellm-omni_2025}: 
An LLM that tokenizes 3D meshes and performs text-based editing by generating mesh tokens that decode into meshes. ShapeLLM-Omni supports geometry-only edits (no texture).
\item LL3M~\citep{lu_ll3m_2025}: An agentic framework that performs 3D editing and generation via Blender script. LL3M converts the input images into into Blender geometric primitives and edits the blender script conditioned on the text instruction. We compare to LL3M qualitatively as their API allows for just 5 queries per day.
\end{itemize} 

\myparit{\textbf{Pipelines using 2D editing:}}
\begin{itemize}
\item \edittrellis: \edittrellis edits the input image with the text instruction using an image editing model, Step1XEdit~\cite{liu_step1x-edit_2025}, and subsequently lifts the edited image to 3D using \trellis~\cite{xiang_structured_2024}, similar to our data engine. 
\item Tailor3D~\cite{qi_tailor3d_2024}: Tailor3D applies image editing on two views of the original 3D asset -- the input image and a back view, then reconstructs the 3D asset. Since it lacks an internal 2D editing module, we employ Step1X-Edit, which is used in our data engine.
\item DGE~\cite{chen_dge_2024}: DGE takes in the pre-edit 3D Gaussian splats ~\cite{kerbl20233d}, and performs multi-view image editing with InstructPix2Pix~\cite{brooks2023instructpix2pix} to optimize the edited Gaussian splats.
\end{itemize}

\mypar{Metrics.}
To thoroughly assess editing quality, we evaluate both geometry and texture. Although there can be multiple correct answers for an editing task, our benchmark is curated to be relatively unambiguous, making alignment with ground truth a reasonable proxy for correctness.

\begin{itemize}
\item Geometry: we compare the prediction to ground truth using Chamfer Distance (lower better) and F1 score (higher better) of $10{,}000$ points sampled on the mesh surface. F1 is evaluated with a threshold of 0.05.
\item Texture: we render out front, back, left, right, up, and down views of the prediction and ground truth assets, and calculate the mean LPIPS~\cite{zhang2018unreasonable} across six views. A lower LPIPS means better alignment with ground truth.
\end{itemize}
Our method preserves orientation and scale, whereas \edittrellis, LL3M, and ShapeLLM-Omni can return objects with mismatched pose or size. To accommodate these baselines, we align their outputs to the ground truth using Iterative Closest Point~\citep{besl1992method} before evaluation.

\subsection{Qualitative Results}

\cref{fig:qualitative} shows qualitative predictions on \bench. \method shows consistent advantage compared to baselines in both faithfulness to the text prompt and consistency with the original generation (pre-edit). 
A key aspect of text steerability in 3D is precise 3D localization of the desired changes. \method shows strong localization capability -- e.g., editing only the lampshade or the chair’s hollow backrest -- while leaving other parts untouched.

Feedforward baselines -- ShapeLLM and LL3M -- struggle to match both the input image and the edit instruction. ShapeLLM often yields broken geometry (\eg, crab, lamp) or reproduces the input asset while ignoring the edit. LL3M, which operates on simplified Blender primitives, fails on cases like crab and cup, producing shapes inconsistent with the image. These issues underscore the difficulty of aligning image, text, and 3D in a single pass. In contrast, \method delivers high-quality 3D edits faithful to both the the prompt and the unsteered (pre-edit) generation -- without any intermediate image editing.

2D–3D pipelines that rely on image editing often suffer from inconsistencies. These inconsistencies arise because small deviations in the edited images can lead to large variations in the reconstructed 3D assets, even beyond the intended local edit. This is evident in the cup example (last row), where \edittrellis reconstructs a thinner, non-symmetric cup -- a side effect of the inconsistency between the reconstructions of post-edit image vs. the original image.
In contrast, \method enables editing by steering its internal diffusion representation; as a result, variations across noise samples are minor, and the edited outputs remain consistent with the original 3D asset. 
Tailor3D exhibits degraded geometry and texture due to mismatches between its edited front and back views. DGE, which performs multiview 2D editing, suffers from analogous inconsistencies, producing floaters and artifacts during Gaussian optimization.

Furthermore, due to \method's 3D nature, it is able to apply edits that are not visible in the front view, as shown in \cref{fig:qualitative} (c). 2D-editing-based pipelines like \edittrellis fails because the 2D editing model cannot apply such edits. This results in hallucination (the multiple tires example), or failure to edit (could not add a cargo door).

\method can also edit 3D assets reconstructed from in-the-wild iPhone or online photos. We provide such examples, as long as more qualitative examples on \method, in the supplementary material.

\subsection{Quantitative Results}


\begin{table}[t!]
\begin{center}
\resizebox{\linewidth}{!}{
\begin{tabular}
{l|ccc|ccc}
\toprule
 & \multicolumn{3}{c|}{Seen Assets, Unseen Edits} & \multicolumn{3}{c}{Unseen Assets} \\
\midrule
  & LPIPS ($\downarrow$)  & Chamfer ($\downarrow$) & F1 ($\uparrow$) & LPIPS  ($\downarrow$) & Chamfer ($\downarrow$) &F1 ($\uparrow$) \\
\midrule
Tailor3D &  0.246 & 0.134 & 0.158 & 0.266 & 0.173 & 0.114 \\
\edittrellis & 0.192 & 0.133 & 0.189 & 0.219 & 0.158 & 0.168 \\
 DGE & 0.265 & 0.252 & 0.075 & 0.233  & 0.267 & 0.064 \\
ShapeLLM & 0.227 & 0.141 & 0.158 & 0.250 & 0.161 & 0.161 \\
\method (Ours) & \textbf{0.142} & \textbf{0.096} & \textbf{0.266} & \textbf{0.125} & \textbf{0.071} & \textbf{0.359} \\
\bottomrule
\end{tabular}
} 
\vspace{-3mm}
\end{center}
\caption{Quantitative comparison on texture edits. We evaluate LPIPS for edit correctness, and Chamfer Distance and F1 score for geometry consistency. We evaluate both on seen assets (unseen edits) and unseen assets. \method shows strong advantage on all metrics in all settings, with a 113\% improvement in F1 score, 55\% reduction in Chamfer Distance, and 43\% reduction in LPIPS.}
\vspace{-4mm}
\label{table:texture}
\end{table}

\begin{figure*}[!htb]
    \centering

    \resizebox{1.8\columnwidth}{!}{
    \begin{minipage}[c]{1\columnwidth}
        \centering
        \small
        \renewcommand{\arraystretch}{0.85}
        
        \begin{tabular}{l|c}
        \toprule
        \method & ``no edit'' failure (\%) \\
        \midrule
        w/o DPO & 18.67 \\
        w/ DPO & \textbf{10.67} \\
        \bottomrule
        \end{tabular}
        \vspace{1mm}
        \\\textbf{(a)}\\

        \vspace{0.7em}

        \begin{tabular}{l|ccc}
        \toprule
          & LPIPS ($\downarrow$) & Chamfer ($\downarrow$) & F1 ($\uparrow$) \\
        \midrule
        No ControlNet & 0.238 & 0.150 & 0.177 \\
        No Data filtering & 0.213 & 0.114 & 0.244 \\
        Current & \textbf{0.1985} & \textbf{0.089} & \textbf{0.2527} \\
        \bottomrule
        \end{tabular}
        \vspace{1mm}
        \\\textbf{(b)}\\

    \end{minipage}
    \hspace{0em} 
    \begin{minipage}[c]{1\columnwidth}
        \centering
        \vspace{0.2em}
        \includegraphics[width=1\columnwidth]{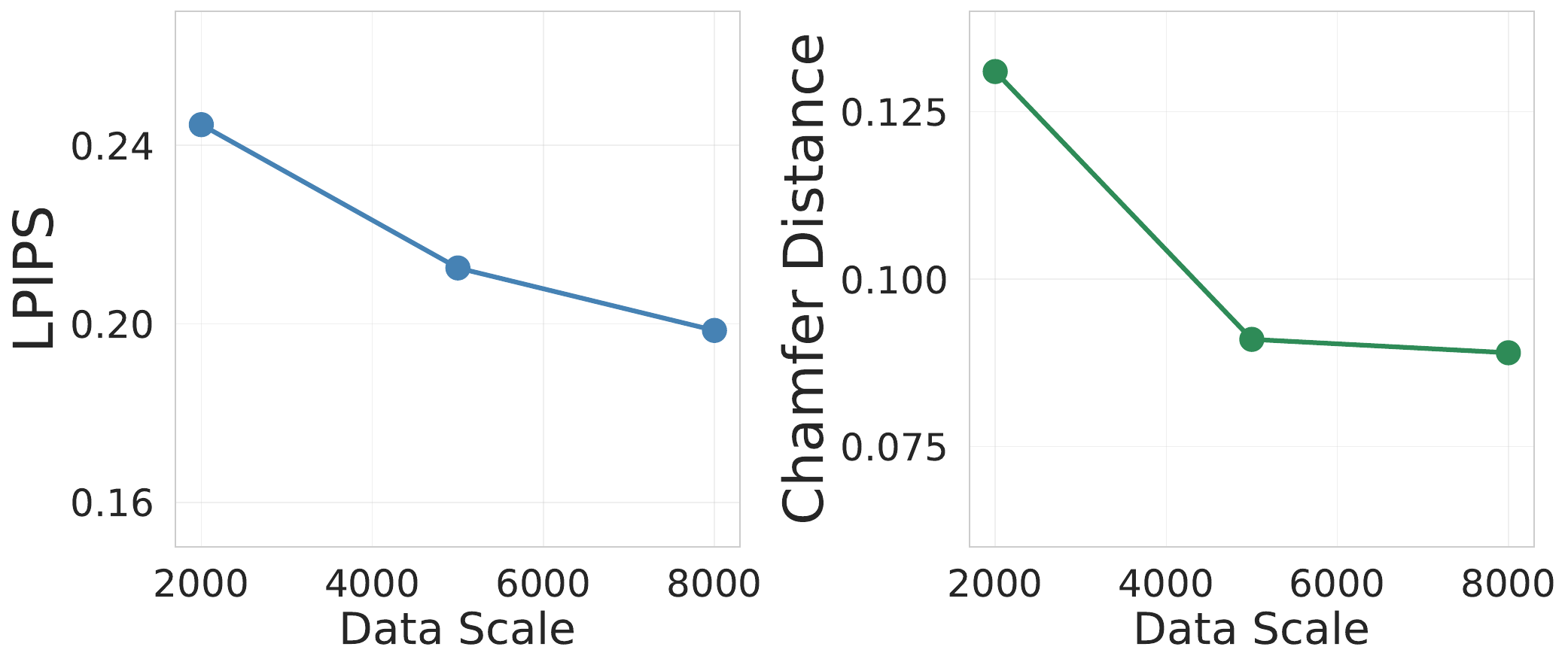}
        \vspace{-1mm}
        \textbf{(c)}
    \end{minipage}
    } 
    \vspace{-3mm}

    \caption{Ablations and scaling analysis of \method. (a): DPO effectively alleviates the failure mode of ``no edit" predictions. (b): Ablations on the ControlNet architecture and data filtering. ``No ControlNet" corresponds to simply adding text as another conditioning signal and finetuning the base model. Both the ControlNet design and data filtering improve all metrics. (c): We see a steady improvement on both LPIPS (lower better) and Chamfer Distance as the training data size increases. Reported metrics are evaluated on the unseen addition set.}
    \label{fig:ablation_combined}
    \vspace{-3mm}
\end{figure*}

We use separate metrics for texture and geometry for a holistic evaluation. We also perform human evaluation in the supplementary material.

\mypar{Texture.}
We evaluate texture editing on the texture subset of \bench in \cref{table:texture}. We measure both appearance metrics, LPIPS, and geometry metrics, Chamfer Distance and F1 score. Since the edits are texture-based, object geometry should not be changed in the output, which is captured by our geometry metrics. 
A predicted asset with perfect geometry consistency should yield a Chamfer Distance of close to 0 and a F1 score of close to 1.
To assess generalization, we evaluate both on seen assets with unseen editing instructions, as well as on unseen assets.

From \cref{table:texture} we observe that \method shows a $43\%$ reduction in LPIPS compared to the second best method (\edittrellis), demonstrating better alignment with ground truth in appearance. Our method also obtains a $55\%$ lower Chamfer Distance and $113\%$ improvement in F1 score, maintaining better geometry consistency.
These results align with the qualitative advantage seen from \cref{fig:qualitative}.

\mypar{Geometry.}
For geometry edits, we evaluate on removal and addition of object parts, each on both seen assets (unseen edits) and unseen assets. LPIPS alignment with ground truth measures both consistency in texture and correctness in geometry, since it measures the 2D per-view alignment with ground truth. Chamfer Distance and F1 score measure the geometry editing correctness.

As seen in \cref{table:geometry}, \method shows strong performance advantage compared to all baselines on all metrics. For removal edits, we achieve a $63\%$ reduction in Chamfer distance and a $64\%$ higher F1 score.
For addition, we achieve a $41\%$ reduction in Chamfer Distance and a $38\%$ higher F1 compared to the second best method. \method also obtains lower LPIPS across all settings, demonstrating better holistic alignment with the ground truth.

\begin{figure}[t]
\vspace{-2mm}
\begin{center}
\centerline{\includegraphics[width=0.7\columnwidth]{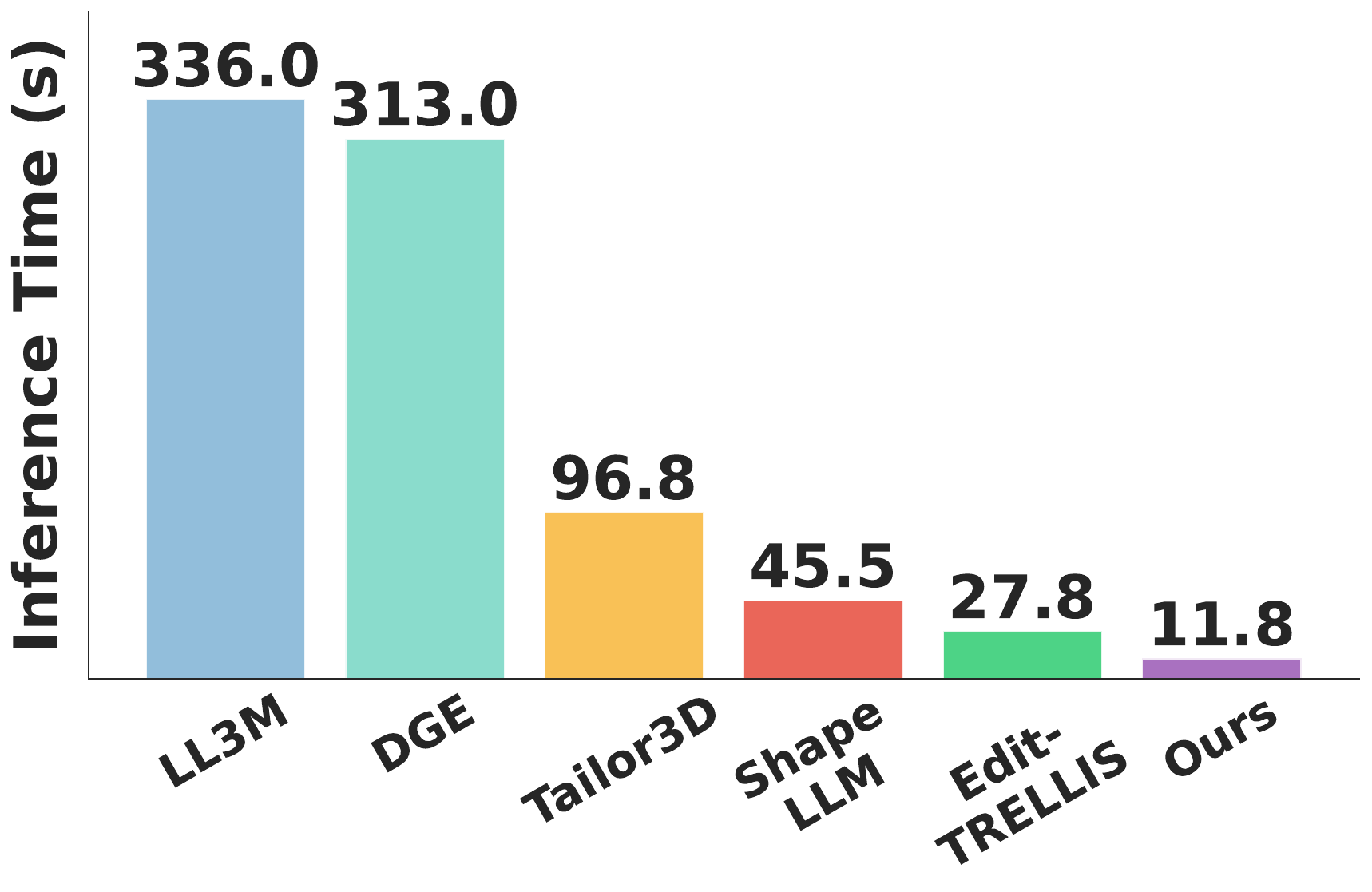}}
\vspace{-2mm}
\caption{Inference time comparison. \method is the fastest -- $2.4 \times$ to $28.5 \times$ faster than competing methods.}
\vspace{-6mm}
\label{fig:time}
\end{center}
\end{figure}

\mypar{Inference Time.}
As seen in \cref{fig:time}, \method is the fastest -- $2.4 \times$ to $28.5 \times$ faster than competing methods. 2D-editing-based pipelines are slow due to rendering, invocation(s) of 2D editing models, and 3D reconstruction. For the feedforward baselines, LL3M is slow due to its agentic design. ShapeLLM is slow due to mesh decoding and an additional texturing step. The reported 11.8s for \method is the total time of both stages, each running 25 steps for the flow model, with classifier-free guidance.

\subsection{Ablations and Analyses}

\begin{figure}[t]
\begin{center}
\centerline{\includegraphics[width=\columnwidth]{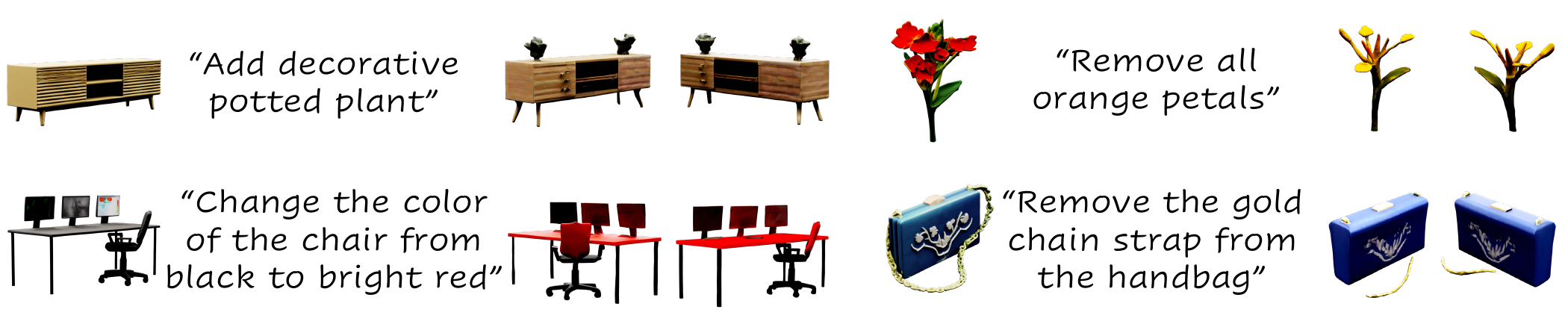}}
\caption{Limitations of \method. For complex edits, \method may yield inconsistency (``add potted plant", ``remove orange petals", edit leaking (``change the chair red"), or partial edits (``remove gold chain").}
\label{fig:limitations}
\end{center}
\vspace{-12mm}
\end{figure}

\mypar{DPO.} Flow-matching supervised learning yields a conservative model that sometimes predicts ``no edit" results. To combat this, we perform a second stage of training using DPO \citep{rafailov_direct_2023} with the original 3D assets as negative examples. \cref{fig:ablation_combined}(a) shows a $8\%$ (absolute) reduction of this failure mode.

\mypar{ControlNet Design.} The alternative architecture design to our method is directly passing text as an additional conditioning signal, and fine-tuning the full \trellis model, instantiated as a DiT \citep{peebles_scalable_2023}.
As shown in \cref{fig:ablation_combined} (b), this alternative results in much worse performance on all metrics, validating the importance of ControlNet.

\mypar{Data Filtering.} Data filtering ``sharpens" the distribution around specific edit types. For example, for addition edits, data filtering removes edit pairs with non-obvious geometry change (via voxel mean absolute error thresholding and texture-keyword filtering), only keeping $41\%$ of the original data. \cref{fig:ablation_combined}(b) shows the effectiveness of filtering on all metrics.


\mypar{Scaling Analysis.}
\cref{fig:ablation_combined}(c) shows continued improvement in model performance as we scale up training data, which corroborates the importance of our data engine.

\mypar{Limitations. }
When the editing instructions are complex, \method exhibits limitations of inconsistency on unedited parts, edit leaking, and partial edits. \cref{fig:limitations} shows some examples.

\section{Conclusions}
We present \method, a method that enables feedforward 3D editing by adding text steering capability to image-to-3D generative models. \method uses an architecture inspired by ControlNet to enable direct text steering via a forward pass. \method works on general objects with complex editing prompts. Compared to existing methods, \method demonstrates better instruction following and improved consistency with original 3D assets, while being $2.4\times$ to $28.5\times$ faster. 
Beyond unlocking a new capability, \method also answers a more general research question: given a pretrained generative model, is it possible to add steering capability via another modality (\eg, text)? In the case of 3D generation, \method shows that this is possible with just less than $100k$ data. We hope that by sharing our recipe, we can inspire similar explorations in other domains.
\method is a general framework that can keep improving with better base models and data generation techniques. We show a scalable path towards general 3D editing and cross-modality steering for generative models.
\section{Acknowledgments}
We would like to thank Xiang Li for discussions on DPO. We also thank Bowen Zhang, Bingliang Zhang, Wenda Chu, Sasha Sax, and Jiacheng Liu for general discussions. Ziqi Ma is supported by the Kortschak scholarship. This project is funded in part by NSF \#2505096, and CAST. 
{
    \small
    \bibliographystyle{ieeenat_fullname}
    \bibliography{main}

@String(ICCV= {Int. Conf. Comput. Vis.})

@String(TOG= {ACM Trans. Graph.})

@String(ICCV  = {ICCV})

@String(TOG   = {ACM TOG})

@article{ye_shapellm-omni_2025,
  title={ShapeLLM-Omni: A Native Multimodal LLM for 3D Generation and Understanding},
  author={Ye, Junliang and Wang, Zhengyi and Zhao, Ruowen and Xie, Shenghao and Zhu, Jun},
  journal={arXiv preprint arXiv:2506.01853},
  year={2025}
}

@article{lu_ll3m_2025,
  title={LL3M: Large Language 3D Modelers},
  author={Lu, Sining and Chen, Guan and Dinh, Nam Anh and Lang, Itai and Holtzman, Ari and Hanocka, Rana},
  journal={arXiv preprint arXiv:2508.08228},
  year={2025}
}

@article{qi_tailor3d_2024,
  title={Tailor3d: Customized 3d assets editing and generation with dual-side images},
  author={Qi, Zhangyang and Yang, Yunhan and Zhang, Mengchen and Xing, Long and Wu, Xiaoyang and Wu, Tong and Lin, Dahua and Liu, Xihui and Wang, Jiaqi and Zhao, Hengshuang},
  journal={arXiv preprint arXiv:2407.06191},
  year={2024}
}

@article{liu_step1x-edit_2025,
  title={Step1x-edit: A practical framework for general image editing},
  author={Liu, Shiyu and Han, Yucheng and Xing, Peng and Yin, Fukun and Wang, Rui and Cheng, Wei and Liao, Jiaqi and Wang, Yingming and Fu, Honghao and Han, Chunrui and others},
  journal={arXiv preprint arXiv:2504.17761},
  year={2025}
}

@article{zhao_hunyuan3d_2025,
  title={Hunyuan3d 2.0: Scaling diffusion models for high resolution textured 3d assets generation},
  author={Zhao, Zibo and Lai, Zeqiang and Lin, Qingxiang and Zhao, Yunfei and Liu, Haolin and Yang, Shuhui and Feng, Yifei and Yang, Mingxin and Zhang, Sheng and Yang, Xianghui and others},
  journal={arXiv preprint arXiv:2501.12202},
  year={2025}
}

@article{hong_lrm_2024,
  title={Lrm: Large reconstruction model for single image to 3d},
  author={Hong, Yicong and Zhang, Kai and Gu, Jiuxiang and Bi, Sai and Zhou, Yang and Liu, Difan and Liu, Feng and Sunkavalli, Kalyan and Bui, Trung and Tan, Hao},
  journal={arXiv preprint arXiv:2311.04400},
  year={2023}
}

@article{rafailov_direct_2023,
  title={Direct preference optimization: Your language model is secretly a reward model},
  author={Rafailov, Rafael and Sharma, Archit and Mitchell, Eric and Manning, Christopher D and Ermon, Stefano and Finn, Chelsea},
  journal={Advances in neural information processing systems},
  volume={36},
  pages={53728--53741},
  year={2023}
}

@inproceedings{sella_vox-e_2023,
  title={Vox-e: Text-guided voxel editing of 3d objects},
  author={Sella, Etai and Fiebelman, Gal and Hedman, Peter and Averbuch-Elor, Hadar},
  booktitle={Proceedings of the IEEE/CVF international conference on computer vision},
  pages={430--440},
  year={2023}
}

@article{wu_unilat3d_2025,
  title={UniLat3D: Geometry-Appearance Unified Latents for Single-Stage 3D Generation},
  author={Wu, Guanjun and Fang, Jiemin and Yang, Chen and Li, Sikuang and Yi, Taoran and Lu, Jia and Zhou, Zanwei and Cen, Jiazhong and Xie, Lingxi and Zhang, Xiaopeng and others},
  journal={arXiv preprint arXiv:2509.25079},
  year={2025}
}

@article{chen_generic_2024,
  title={Generic 3d diffusion adapter using controlled multi-view editing},
  author={Chen, Hansheng and Shi, Ruoxi and Liu, Yulin and Shen, Bokui and Gu, Jiayuan and Wetzstein, Gordon and Su, Hao and Guibas, Leonidas},
  journal={arXiv preprint arXiv:2403.12032},
  year={2024}
}

@inproceedings{erkoc_preditor3d_2024,
  title={Preditor3d: Fast and precise 3d shape editing},
  author={Erko{\c{c}}, Ziya and G{\"u}meli, Can and Wang, Chaoyang and Nie{\ss}ner, Matthias and Dai, Angela and Wonka, Peter and Lee, Hsin-Ying and Zhuang, Peiye},
  booktitle={Proceedings of the Computer Vision and Pattern Recognition Conference},
  pages={640--649},
  year={2025}
}

@inproceedings{wu_gaussctrl_2024,
  title={Gaussctrl: Multi-view consistent text-driven 3d gaussian splatting editing},
  author={Wu, Jing and Bian, Jia-Wang and Li, Xinghui and Wang, Guangrun and Reid, Ian and Torr, Philip and Prisacariu, Victor Adrian},
  booktitle={European Conference on Computer Vision},
  pages={55--71},
  year={2024},
  organization={Springer}
}

@inproceedings{khalid_latenteditor_2024,
  title={Latenteditor: Text driven local editing of 3d scenes},
  author={Khalid, Umar and Iqbal, Hasan and Karim, Nazmul and Tayyab, Muhammad and Hua, Jing and Chen, Chen},
  booktitle={European Conference on Computer Vision},
  pages={364--380},
  year={2024},
  organization={Springer}
}

@inproceedings{chen_consistdreamer_2024,
  title={Consistdreamer: 3d-consistent 2d diffusion for high-fidelity scene editing},
  author={Chen, Jun-Kun and Bulo, Samuel Rota and M{\"u}ller, Norman and Porzi, Lorenzo and Kontschieder, Peter and Wang, Yu-Xiong},
  booktitle={Proceedings of the IEEE/CVF Conference on Computer Vision and Pattern Recognition},
  pages={21071--21080},
  year={2024}
}

@inproceedings{hong_perturb-and-revise_2025,
  title={Perturb-and-Revise: Flexible 3D Editing with Generative Trajectories},
  author={Hong, Susung and Karras, Johanna and Martin-Brualla, Ricardo and Kemelmacher-Shlizerman, Ira},
  booktitle={Proceedings of the Computer Vision and Pattern Recognition Conference},
  pages={16293--16303},
  year={2025}
}

@inproceedings{gao_3d_2025,
  title={3d mesh editing using masked lrms},
  author={Gao, Will and Wang, Dilin and Fan, Yuchen and Bozic, Aljaz and Stuyck, Tuur and Li, Zhengqin and Dong, Zhao and Ranjan, Rakesh and Sarafianos, Nikolaos},
  booktitle={Proceedings of the IEEE/CVF International Conference on Computer Vision},
  pages={7154--7165},
  year={2025}
}

@inproceedings{karim_free-editor_2024,
  title={Free-editor: zero-shot text-driven 3D scene editing},
  author={Karim, Nazmul and Iqbal, Hasan and Khalid, Umar and Chen, Chen and Hua, Jing},
  booktitle={European Conference on Computer Vision},
  pages={436--453},
  year={2024},
  organization={Springer}
}

@article{poole_dreamfusion_2022,
  title={Dreamfusion: Text-to-3d using 2d diffusion},
  author={Poole, Ben and Jain, Ajay and Barron, Jonathan T and Mildenhall, Ben},
  journal={arXiv preprint arXiv:2209.14988},
  year={2022}
}

@article{zhuang_tip-editor_2024,
  title={Tip-editor: An accurate 3d editor following both text-prompts and image-prompts},
  author={Zhuang, Jingyu and Kang, Di and Cao, Yan-Pei and Li, Guanbin and Lin, Liang and Shan, Ying},
  journal={ACM Transactions on Graphics (TOG)},
  volume={43},
  number={4},
  pages={1--12},
  year={2024},
  publisher={ACM New York, NY, USA}
}

@article{parelli_3d-latte_2025,
  title={3D-LATTE: Latent Space 3D Editing from Textual Instructions},
  author={Parelli, Maria and Oechsle, Michael and Niemeyer, Michael and Tombari, Federico and Geiger, Andreas},
  journal={arXiv preprint arXiv:2509.00269},
  year={2025}
}

@inproceedings{huang_blenderalchemy_2024,
  title={Blenderalchemy: Editing 3d graphics with vision-language models},
  author={Huang, Ian and Yang, Guandao and Guibas, Leonidas},
  booktitle={European Conference on Computer Vision},
  pages={297--314},
  year={2024},
  organization={Springer}
}

@article{yang_hunyuan3d_2025,
  title={Hunyuan3d 1.0: A unified framework for text-to-3d and image-to-3d generation},
  author={Yang, Xianghui and Shi, Huiwen and Zhang, Bowen and Yang, Fan and Wang, Jiacheng and Zhao, Hongxu and Liu, Xinhai and Wang, Xinzhou and Lin, Qingxiang and Yu, Jiaao and others},
  journal={arXiv preprint arXiv:2411.02293},
  year={2024}
}

@article{deitke_objaverse-xl_2023,
  title={Objaverse-xl: A universe of 10m+ 3d objects},
  author={Deitke, Matt and Liu, Ruoshi and Wallingford, Matthew and Ngo, Huong and Michel, Oscar and Kusupati, Aditya and Fan, Alan and Laforte, Christian and Voleti, Vikram and Gadre, Samir Yitzhak and others},
  journal={Advances in Neural Information Processing Systems},
  volume={36},
  pages={35799--35813},
  year={2023}
}

@article{ho_classifier-free_2022,
  title={Classifier-free diffusion guidance},
  author={Ho, Jonathan and Salimans, Tim},
  journal={arXiv preprint arXiv:2207.12598},
  year={2022}
}

@inproceedings{hessel_clipscore_2022,
  title={Clipscore: A reference-free evaluation metric for image captioning},
  author={Hessel, Jack and Holtzman, Ari and Forbes, Maxwell and Le Bras, Ronan and Choi, Yejin},
  booktitle={Proceedings of the 2021 conference on empirical methods in natural language processing},
  pages={7514--7528},
  year={2021}
}

@article{kirstain_pick--pic_2023,
  title={Pick-a-pic: An open dataset of user preferences for text-to-image generation},
  author={Kirstain, Yuval and Polyak, Adam and Singer, Uriel and Matiana, Shahbuland and Penna, Joe and Levy, Omer},
  journal={Advances in neural information processing systems},
  volume={36},
  pages={36652--36663},
  year={2023}
}

@inproceedings{duggal_eval3d_2025,
  title={Eval3D: Interpretable and fine-grained evaluation for 3D generation},
  author={Duggal, Shivam and Hu, Yushi and Michel, Oscar and Kembhavi, Aniruddha and Freeman, William T and Smith, Noah A and Krishna, Ranjay and Torralba, Antonio and Farhadi, Ali and Ma, Wei-Chiu},
  booktitle={Proceedings of the Computer Vision and Pattern Recognition Conference},
  pages={13326--13336},
  year={2025}
}

@article{liu2025improving,
  title={Improving video generation with human feedback},
  author={Liu, Jie and Liu, Gongye and Liang, Jiajun and Yuan, Ziyang and Liu, Xiaokun and Zheng, Mingwu and Wu, Xiele and Wang, Qiulin and Qin, Wenyu and Xia, Menghan and others},
  journal={arXiv preprint arXiv:2501.13918},
  year={2025}
}

@inproceedings{chen2024shap,
  title={Shap-editor: Instruction-guided latent 3d editing in seconds},
  author={Chen, Minghao and Xie, Junyu and Laina, Iro and Vedaldi, Andrea},
  booktitle={Proceedings of the IEEE/CVF conference on computer vision and pattern recognition},
  pages={26456--26466},
  year={2024}
}

@inproceedings{lugmayr2023inpainting,
  title={Inpainting using denoising diffusion probabilistic models},
  author={Lugmayr, Andreas and Danelljan, Martin and Romero, Andres and Yu, Fisher and Timofte, Radu and Van Gool, L Repaint},
  booktitle={Proceedings of the IEEE/CVF Conference on Computer Vision and Pattern Recognition},
  pages={11461--11471},
  year={2023}
}

@inproceedings{zhang2018unreasonable,
  title={The unreasonable effectiveness of deep features as a perceptual metric},
  author={Zhang, Richard and Isola, Phillip and Efros, Alexei A and Shechtman, Eli and Wang, Oliver},
  booktitle={Proceedings of the IEEE conference on computer vision and pattern recognition},
  pages={586--595},
  year={2018}
}

@inproceedings{besl1992method,
  title={Method for registration of 3-D shapes},
  author={Besl, Paul J and McKay, Neil D},
  booktitle={Sensor fusion IV: control paradigms and data structures},
  volume={1611},
  pages={586--606},
  year={1992},
  organization={Spie}
}

@inproceedings{barda2025instant3dit,
  title={Instant3dit: Multiview inpainting for fast editing of 3d objects},
  author={Barda, Amir and Gadelha, Matheus and Kim, Vladimir G and Aigerman, Noam and Bermano, Amit H and Groueix, Thibault},
  booktitle={Proceedings of the Computer Vision and Pattern Recognition Conference},
  pages={16273--16282},
  year={2025}
}

@article{kerbl20233d,
  title={3D Gaussian splatting for real-time radiance field rendering.},
  author={Kerbl, Bernhard and Kopanas, Georgios and Leimk{\"u}hler, Thomas and Drettakis, George},
  journal={ACM Trans. Graph.},
  volume={42},
  number={4},
  pages={139--1},
  year={2023}
}

@article{mildenhall2021nerf,
  title={Nerf: Representing scenes as neural radiance fields for view synthesis},
  author={Mildenhall, Ben and Srinivasan, Pratul P and Tancik, Matthew and Barron, Jonathan T and Ramamoorthi, Ravi and Ng, Ren},
  journal={Communications of the ACM},
  volume={65},
  number={1},
  pages={99--106},
  year={2021},
  publisher={ACM New York, NY, USA}
}

@InProceedings{Kim_2025_ICCV,
    author    = {Kim, Hayeon and Jang, Ji Ha and Chun, Se Young},
    title     = {Robust 3D-Masked Part-level Editing in 3D Gaussian Splatting with Regularized Score Distillation Sampling},
    booktitle = {Proceedings of the IEEE/CVF International Conference on Computer Vision (ICCV)},
    month     = {October},
    year      = {2025},
    pages     = {5501-5510}
}

@inproceedings{brooks2023instructpix2pix,
  title={Instructpix2pix: Learning to follow image editing instructions},
  author={Brooks, Tim and Holynski, Aleksander and Efros, Alexei A},
  booktitle={Proceedings of the IEEE/CVF conference on computer vision and pattern recognition},
  pages={18392--18402},
  year={2023}
}

@inproceedings{peebles_scalable_2023,
    address = {Paris, France},
    title = {Scalable {Diffusion} {Models} with {Transformers}},
    copyright = {https://doi.org/10.15223/policy-029},
    isbn = {979-8-3503-0718-4},
    url = {https://ieeexplore.ieee.org/document/10377858/},
    doi = {10.1109/ICCV51070.2023.00387},
    language = {en},
    urldate = {2025-11-14},
    booktitle = {2023 {IEEE}/{CVF} {International} {Conference} on {Computer} {Vision} ({ICCV})},
    publisher = {IEEE},
    author = {Peebles, William and Xie, Saining},
    month = oct,
    year = {2023},
    pages = {4172--4182},
}

@misc{xiang_structured_2024,
    title = {Structured {3D} {Latents} for {Scalable} and {Versatile} {3D} {Generation}},
    url = {http://arxiv.org/abs/2412.01506},
    doi = {10.48550/arXiv.2412.01506},
    urldate = {2025-02-19},
    publisher = {arXiv},
    author = {Xiang, Jianfeng and Lv, Zelong and Xu, Sicheng and Deng, Yu and Wang, Ruicheng and Zhang, Bowen and Chen, Dong and Tong, Xin and Yang, Jiaolong},
    month = dec,
    year = {2024},
    note = {arXiv:2412.01506 [cs]},
}

@inproceedings{chen_dge_2024,
  title={Dge: Direct gaussian 3d editing by consistent multi-view editing},
  author={Chen, Minghao and Laina, Iro and Vedaldi, Andrea},
  booktitle={European Conference on Computer Vision},
  pages={74--92},
  year={2024},
  organization={Springer}
}

@inproceedings{haque_instruct-nerf2nerf_2023,
  title={Instruct-nerf2nerf: Editing 3d scenes with instructions},
  author={Haque, Ayaan and Tancik, Matthew and Efros, Alexei A and Holynski, Aleksander and Kanazawa, Angjoo},
  booktitle={Proceedings of the IEEE/CVF international conference on computer vision},
  pages={19740--19750},
  year={2023}
}

@inproceedings{he_customize_2023,
  title={Customize your nerf: Adaptive source driven 3d scene editing via local-global iterative training},
  author={He, Runze and Huang, Shaofei and Nie, Xuecheng and Hui, Tianrui and Liu, Luoqi and Dai, Jiao and Han, Jizhong and Li, Guanbin and Liu, Si},
  booktitle={Proceedings of the IEEE/CVF conference on computer vision and pattern recognition},
  pages={6966--6975},
  year={2024}
}

@article{fu_dreamsim_2023,
  title={Dreamsim: Learning new dimensions of human visual similarity using synthetic data},
  author={Fu, Stephanie and Tamir, Netanel and Sundaram, Shobhita and Chai, Lucy and Zhang, Richard and Dekel, Tali and Isola, Phillip},
  journal={arXiv preprint arXiv:2306.09344},
  year={2023}
}

@inproceedings{wang_view-consistent_2025,
  title={View-consistent 3d editing with gaussian splatting},
  author={Wang, Yuxuan and Yi, Xuanyu and Wu, Zike and Zhao, Na and Chen, Long and Zhang, Hanwang},
  booktitle={European conference on computer vision},
  pages={404--420},
  year={2024},
  organization={Springer}
}

@inproceedings{zhang_adding_2023,
  title={Adding conditional control to text-to-image diffusion models},
  author={Zhang, Lvmin and Rao, Anyi and Agrawala, Maneesh},
  booktitle={Proceedings of the IEEE/CVF international conference on computer vision},
  pages={3836--3847},
  year={2023}
}

@article{lipman_flow_2023,
  title={Flow matching for generative modeling},
  author={Lipman, Yaron and Chen, Ricky TQ and Ben-Hamu, Heli and Nickel, Maximilian and Le, Matt},
  journal={arXiv preprint arXiv:2210.02747},
  year={2022}
}
}

\clearpage
\appendix
\setcounter{page}{1}
\maketitlesupplementary

\section{Additional Qualitative Examples of \method}
\begin{figure*}[ht]
\begin{center}
\centerline{\includegraphics[width=2\columnwidth]{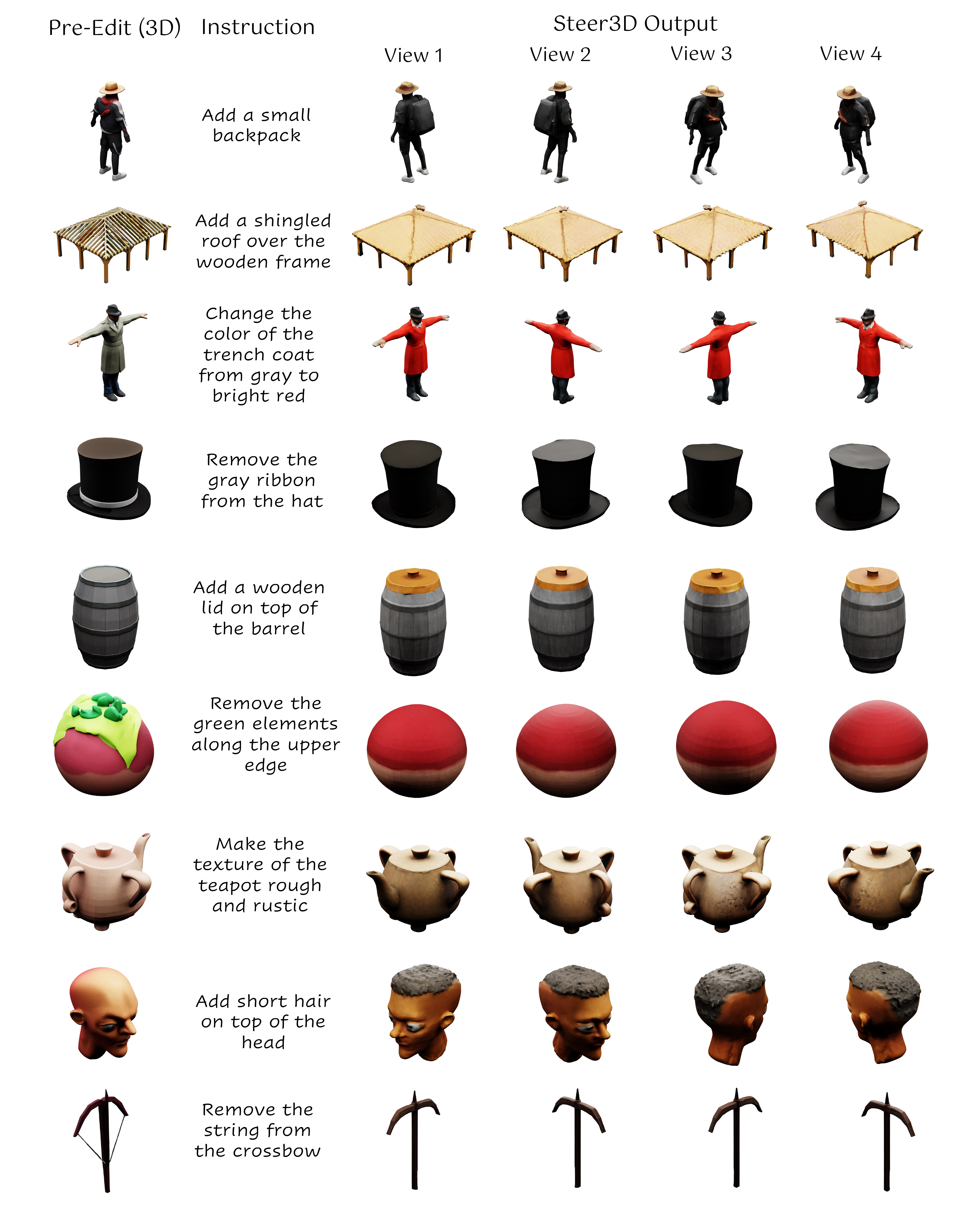}}
\caption{Additional qualitative examples on \bench. Pre-Edit (3D) shows the base image-to-3D model generation that we want to steer. \method output is shown in 4 views.}
\label{fig:qualitative1}
\end{center}
\end{figure*}

\begin{figure*}[ht]
\begin{center}
\centerline{\includegraphics[width=2\columnwidth]{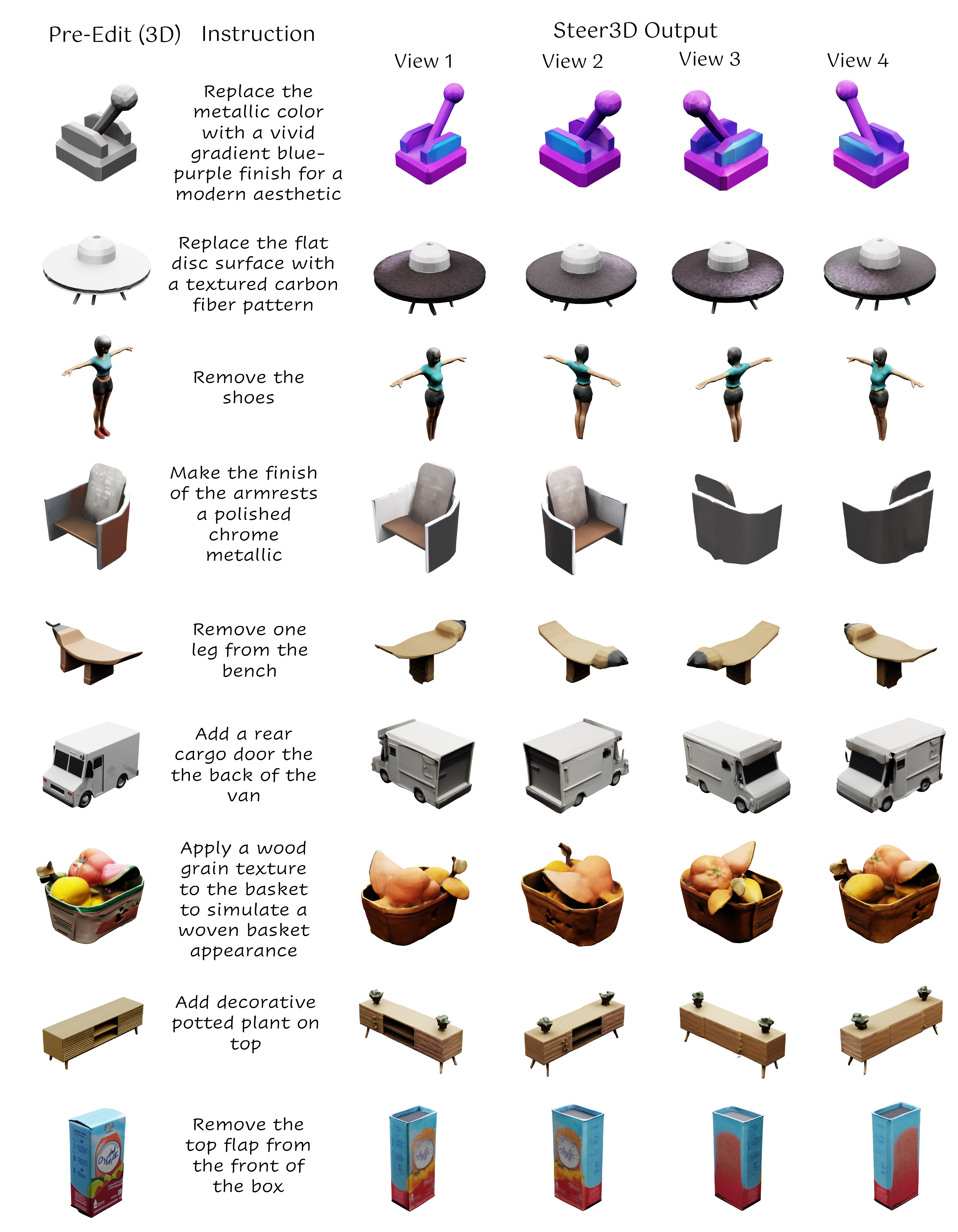}}
\caption{Additional qualitative examples on \bench. Pre-Edit (3D) shows the base image-to-3D model generation that we want to steer. \method output is shown in 4 views.}
\label{fig:qualitative2}
\end{center}
\end{figure*}

\begin{figure*}[ht]
\begin{center}
\centerline{\includegraphics[width=2\columnwidth]{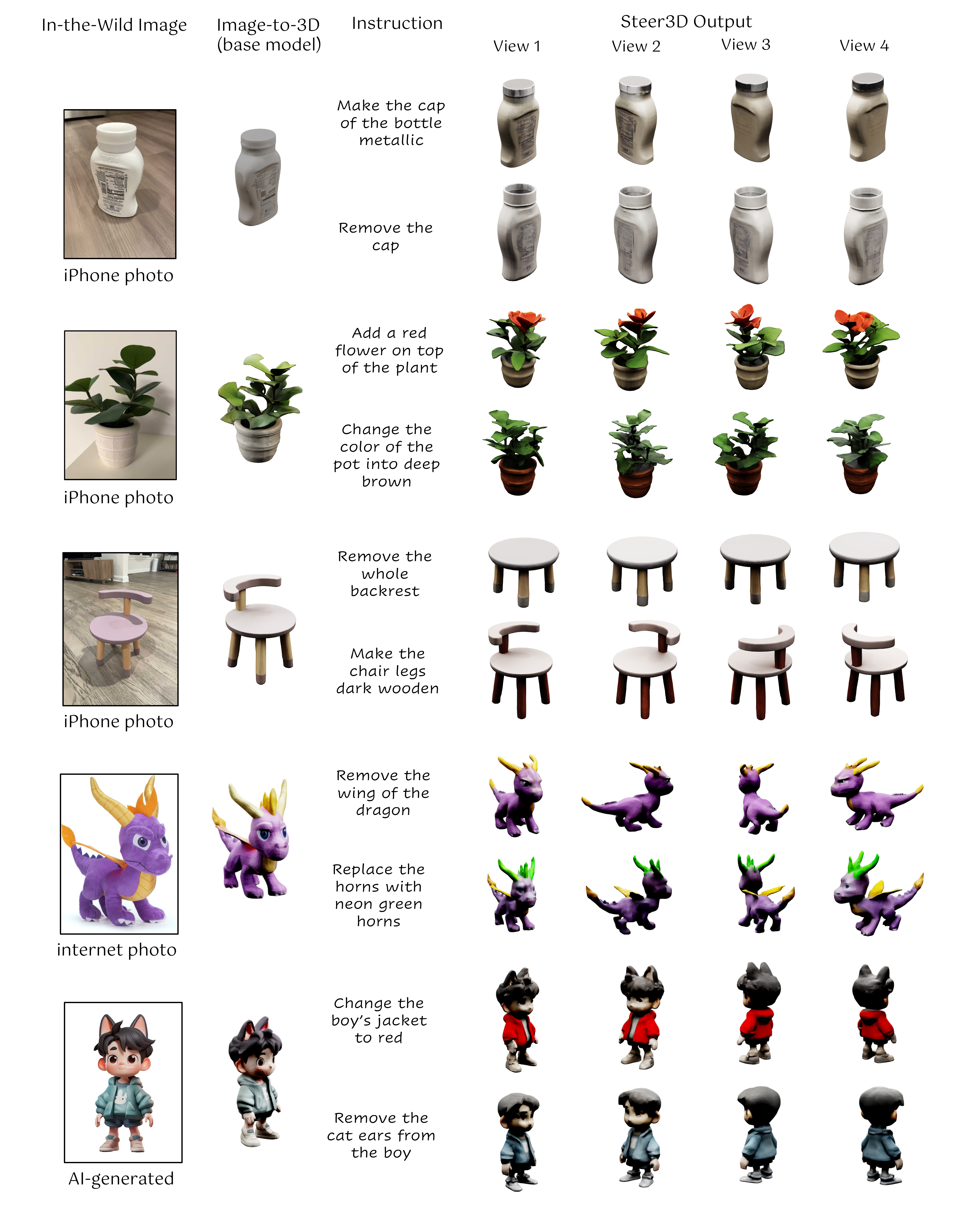}}
\caption{Additional qualitative examples on challenging in-the-wild objects, captured by iPhone photos, online photo, or AI-generated image. Image-to-3D shows the original base model output based on the input image. \method output is shown in 4 views. We show best-of-3 output for in-the-wild evaluation.}
\label{fig:qualitativewild}
\end{center}
\end{figure*}

We show additional qualitative examples both on our benchmark and ``in the wild" -- from objects in iPhone or online photos, or AI-generated images. \cref{fig:qualitative1} and \cref{fig:qualitative2} show additional examples on our benchmark in 4 views, and \cref{fig:qualitativewild} shows examples on ``in-the-wild" objects based on iPhone or online photos, or an AI-generated image. ``In-the-wild" evaluation is challenging: \method is only trained on synthetic data based on Objaverse~\cite{deitke_objaverse-xl_2023} assets, and reconstructed real-world objects are out of distribution. Even in this challenging setting, \method is able to perform edits correctly and preserve consistency with the original asset.

\section{Additional Examples of \bench}
We provide additional examples of \bench in \cref{fig:benchmark} to show that \bench is diverse and representative of edits that a normal user would apply to generated 3D assets.

\section{Human Evaluation}
\begin{figure*}[ht]
\begin{center}
\centerline{\includegraphics[width=2\columnwidth]{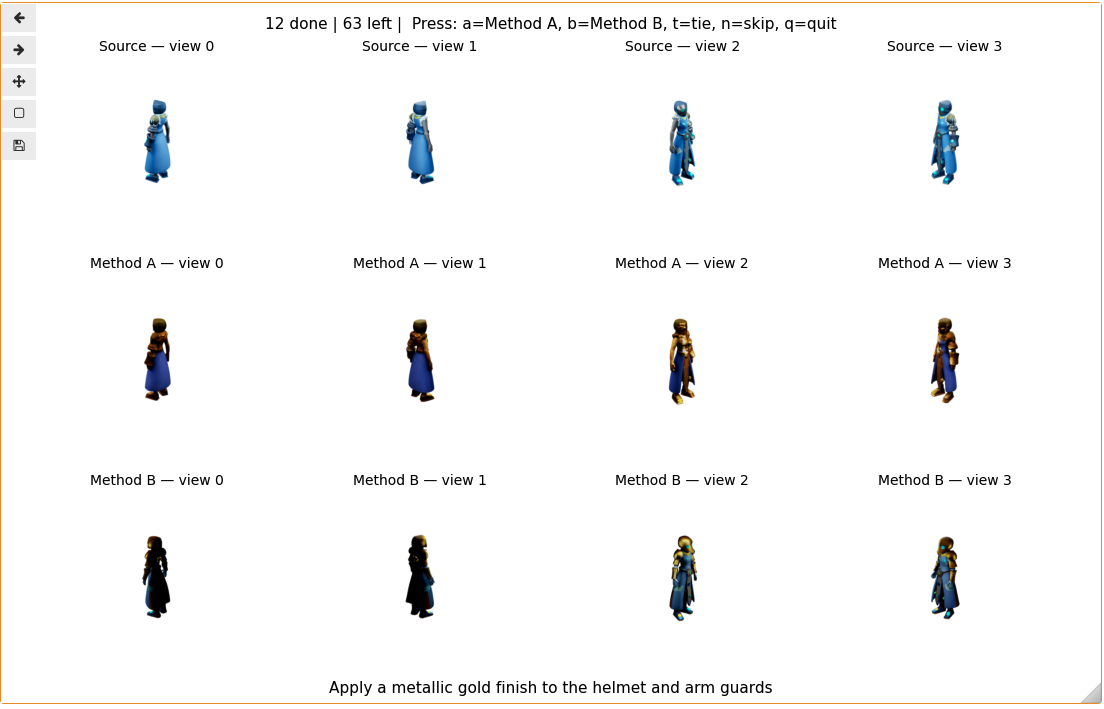}}
\caption{User interface for human evaluation. The top row shows the pre-edit 3D asset rendered in 4 views. The middle and bottom rows are two blinded methods in randomized order per example. The user presses ``A" for method A, ``B" for method B, or ``T for a tie.}
\label{fig:ui}
\end{center}
\end{figure*}

\begin{table*}[ht]
\begin{center}
\resizebox{\linewidth}{!}{
\begin{tabular}
{l|cc|cc|cc|c}
\toprule
Win Rate ($\%$) & \multicolumn{2}{c|}{Removal} & \multicolumn{2}{c|}{Addition} & \multicolumn{2}{c|}{Texture} & Total \\
 \midrule
 & \text{Seen Assets, Unseen Edits} & {Unseen Assets} & \text{Seen Assets, Unseen Edits} & {Unseen Assets} & \text{Seen Assets, Unseen Edits} & {Unseen Assets} & \\
\midrule
\edittrellis  & 44.0 & 24.0 & 37.3 & 48.0 & 33.3 & 25.3 & 32.9\\
\method & \textbf{56.0} & \textbf{76.0} & \textbf{62.7} & \textbf{52.0} & \textbf{66.7} & \textbf{74.7} & \textbf{67.1}\\
\bottomrule
\end{tabular}
} 
\end{center}
\vspace{-5mm}
\caption{Human preference results (win rate) of \edittrellis vs. \method, ranked by 3 independent annotators. \method wins more than $2:1$ compared to \edittrellis.}
\label{table:humaneval}
\end{table*}
We conduct an additional double-blind experiment on \bench between \method and the strongest competing method, \edittrellis, and show results in \cref{table:humaneval}. \method wins over \edittrellis with an over $2:1$ win rate. The experiment UI is shown in \cref{fig:ui}. The top row shows the pre-edit 3D asset rendered in 4 views. The editing instruction is shown at the bottom. The middle and bottom rows are two blinded methods in randomized order per example -- i.e., no method is consistently shown on top or below. The user is instructed to evaluate whether the edited asset not only follows the text instruction but also keeps consistent with the original asset, in shape, appearance (modulo the edit), size, and orientation. The annotators are encouraged to use the ``tie" option sparingly -- less than $2\%$ from the experiment result. When calculating win rates, we do not include tied comparisons. Three annotators conduct the experiment independently, and we report the average of their preference results.

\section{Additional Training Details and Analyses}
\label{sec:additional_ablation}
Section 5.3 of the main paper provides key ablations and scaling analyses. We provide additional analyses in this section.

\begin{figure}[h!]
\begin{center}
\vspace{3mm}
\centerline{\includegraphics[width=0.7\columnwidth]{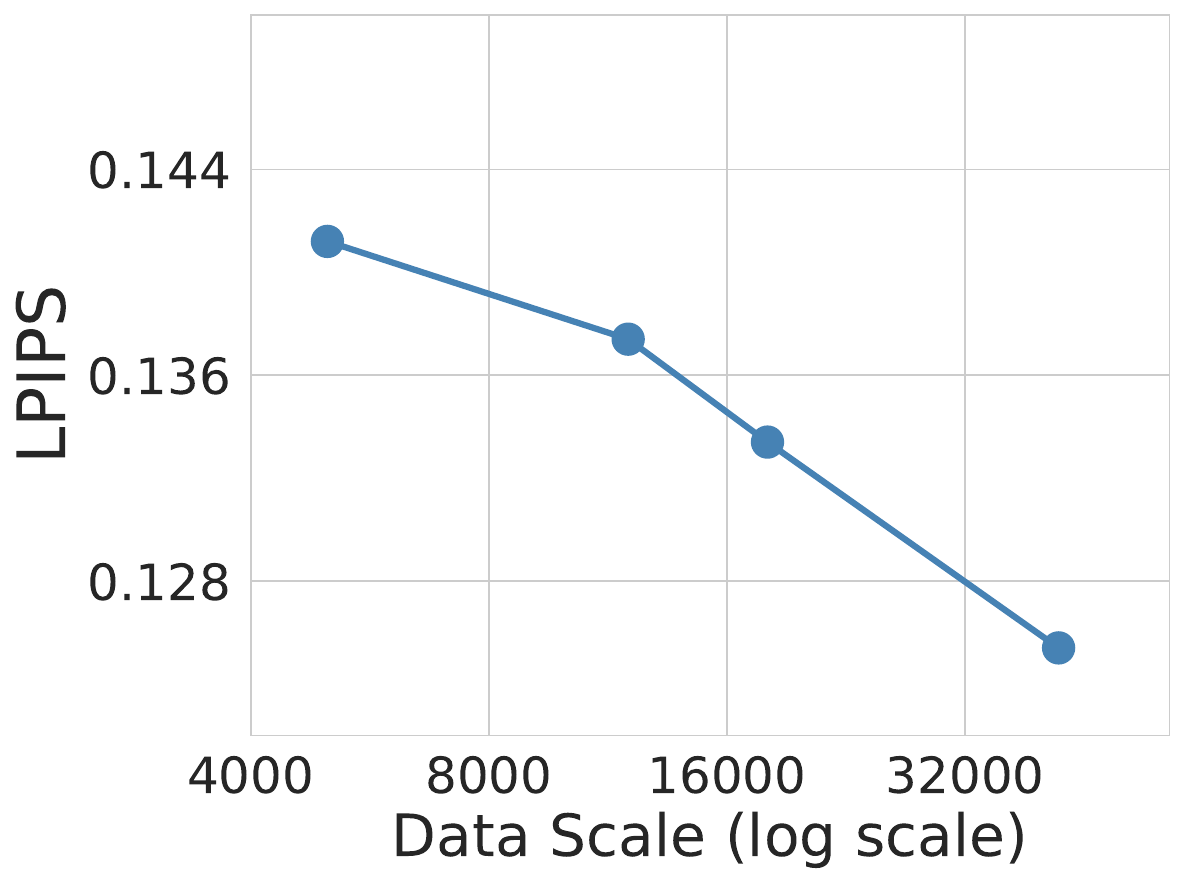}}
\caption{Scaling analysis on texture edits, evaluated on unseen assets. As the training size grows, LPIPS decreases steadily, showing the importance of scaling up. The x axis is plotted on log scale.}
\vspace{-3mm}
\label{fig:scaling_texture}
\end{center}
\end{figure}

\mypar{Additional Scaling Analyses. } We further show the scaling behavior on texture in \cref{fig:scaling_texture}, evaluated by average LPIPS from renderings of the prediction vs. ground truth on unseen assets, as detailed in main paper Section 5. Similar to the scaling trends on the geometry metrics shown in the main paper (Section 5.3), we observe a steady decrease of LPIPS as the training data scales up, which further corroborates the importance of our scalable data engine and training recipe.

\mypar{Classifier-Free Guidance. } In our experiments, we notice that when the dataset size is small, Classifier-Free Guidance~\citep{ho_classifier-free_2022} can have a negative effect, whereas it helps at large scale. One possible explanation for this observation is that CFG requires a strong-enough conditional model, which only emerges when the training scale is large enough.
\begin{table}[h]
\begin{center}
\resizebox{0.8\linewidth}{!}{
\begin{tabular}
{l|c|c}
\toprule
Chamfer Distance ($\downarrow$)  & $8k$-Scale & $13k$-Scale\\
\midrule
without CFG &	0.1206 & 0.1065 \\
with CFG & 0.1423 & 0.1021 \\
\bottomrule
\end{tabular}
} 
\end{center}
\vspace{-5mm}
\caption{The effect of CFG at small vs. large data regime. We evaluate the effect of CFG with varying sizes of training data. CFG hurts performance when the training set is small ($8k$-scale), but improves performance when the training set is large ($13k$-scale). The reported metrics are from the unseen-removal set.}
\label{table:ablation_cfg}
\end{table}

\mypar{Additional Training Details. }
Section 3.3 of the main paper briefly discusses training details. We list additional hyperparameters in \cref{table:hyperparams}. We use bfloat16 during training to improve speed, and use gradient checkpointing for stage 2 and DPO training to save memory.
Feature normalization is applied to the per-voxel latents during training, as in the original TRELLIS~\citep{xiang_structured_2024} paper.
\begin{table}[h]
\begin{center}
\vspace{3mm}
\resizebox{\linewidth}{!}{
\begin{tabular}
{l|c|c|c}
\toprule
 & Stage 1 & Stage 2 & DPO (texture model) \\
\midrule
Effective batch size &	12 & 48 & 12 \\
Gradient accumulation step & 2 & 1 & 2 \\
Learning rate (AdamW) & 2e-5 & 5e-5 & 1e-6 \\
Training precision & bfloat16 & bfloat16 & bfloat16 \\
\shortstack[l]{Timestep sampling \\ (LogitNormal, mean;std)} & 1.0;1.8 & 1.0;1.0 & 1.0;1.0 \\
Gradient checkpointing & No & Yes & Yes \\
CFG-$p_\text{uncond}$ & 0.0 & 0.2 & 0.0 \\
DPO-$\beta$ & - & - & 0.2 \\
DPO-supervised loss weight ($\alpha$) & - & - & 1.0 \\
\bottomrule
\end{tabular}
} 
\end{center}
\vspace{-5mm}
\caption{Additional training hyperparameters.}
\label{table:hyperparams}
\end{table}

Furthermore, because our data engine leverages Hunyuan~\citep{zhao_hunyuan3d_2025} whereas our base model uses TRELLIS~\citep{xiang_structured_2024}, sometimes the base model prediction might have orientation or size differences from the edited version. Naive ControlNet training puts the additional burden of aligning orientation or size to the control branch, which makes the supervision signal of text-to-3D mapping less clear. To combat this, we first perform supervised finetuning of TRELLIS on Hunyuan outputs for the image-to-3D task, so that the base model prediction and the edited ``ground truth" is generally aligned (in size and orientation), except for the edit.

\begin{figure*}[ht]
\begin{center}
\centerline{\includegraphics[width=2\columnwidth]{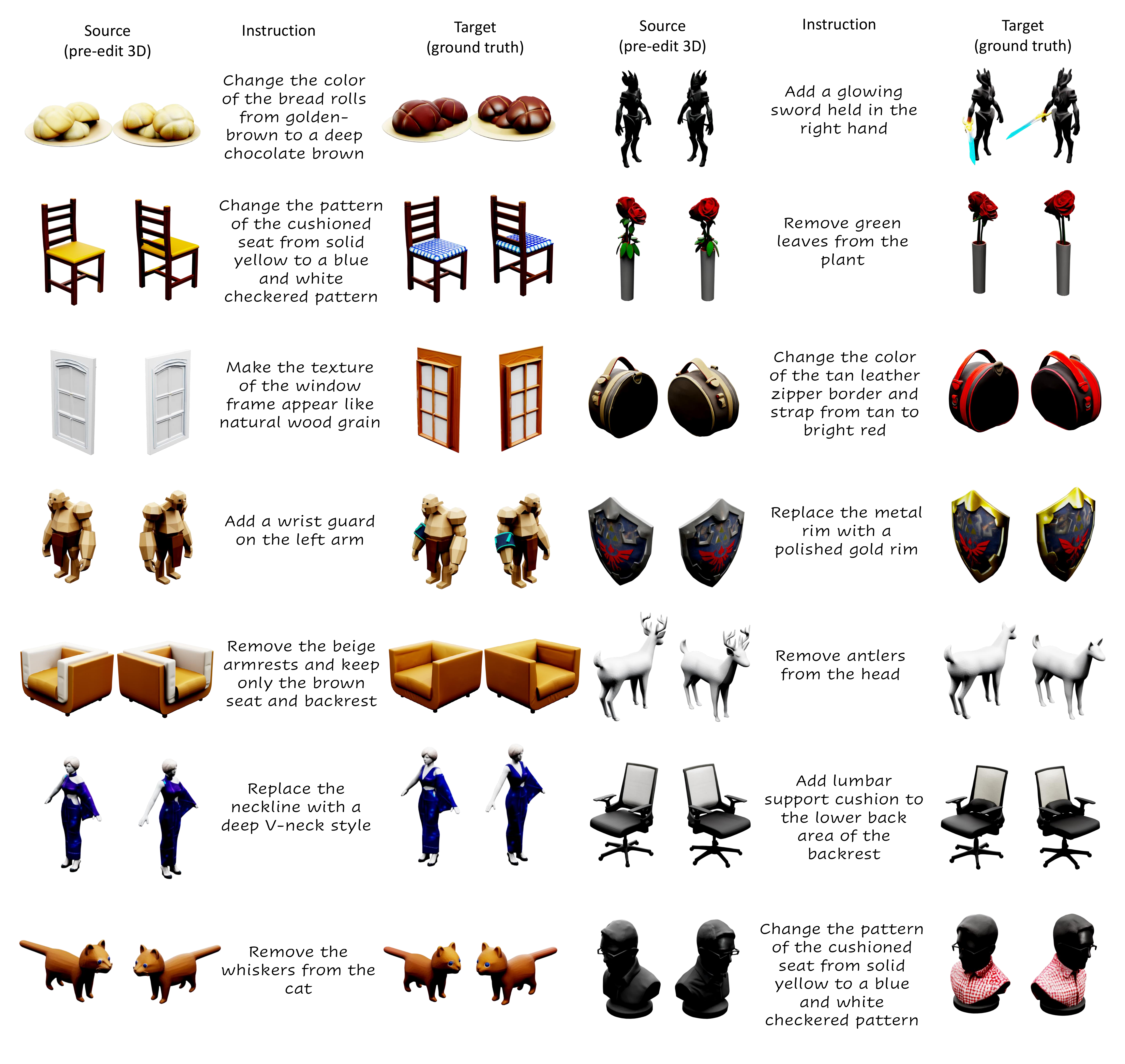}}
\caption{We provide additional examples from \bench, which shows that \bench is diverse.}
\label{fig:benchmark}
\end{center}
\end{figure*}

\section{Additional Data Engine Details}
We provide additional details about the data engine. As mentioned in Section 3.2, we use a two-stage filter to improve data quality.

\mypar{LLM-Based Filtering of Editing Correctness. }
We use LLM-based filtering to ensure that the editing is performed correctly. Directly asking a VLM whether the edit is carried out correctly results in hallucinations. We hypothesis that this is because distinguishing localized edits requires fine-grained visual perception capabilities, which current VLMs struggle with.

\begin{figure*}[ht]
\begin{center}
\centerline{\includegraphics[width=2\columnwidth]{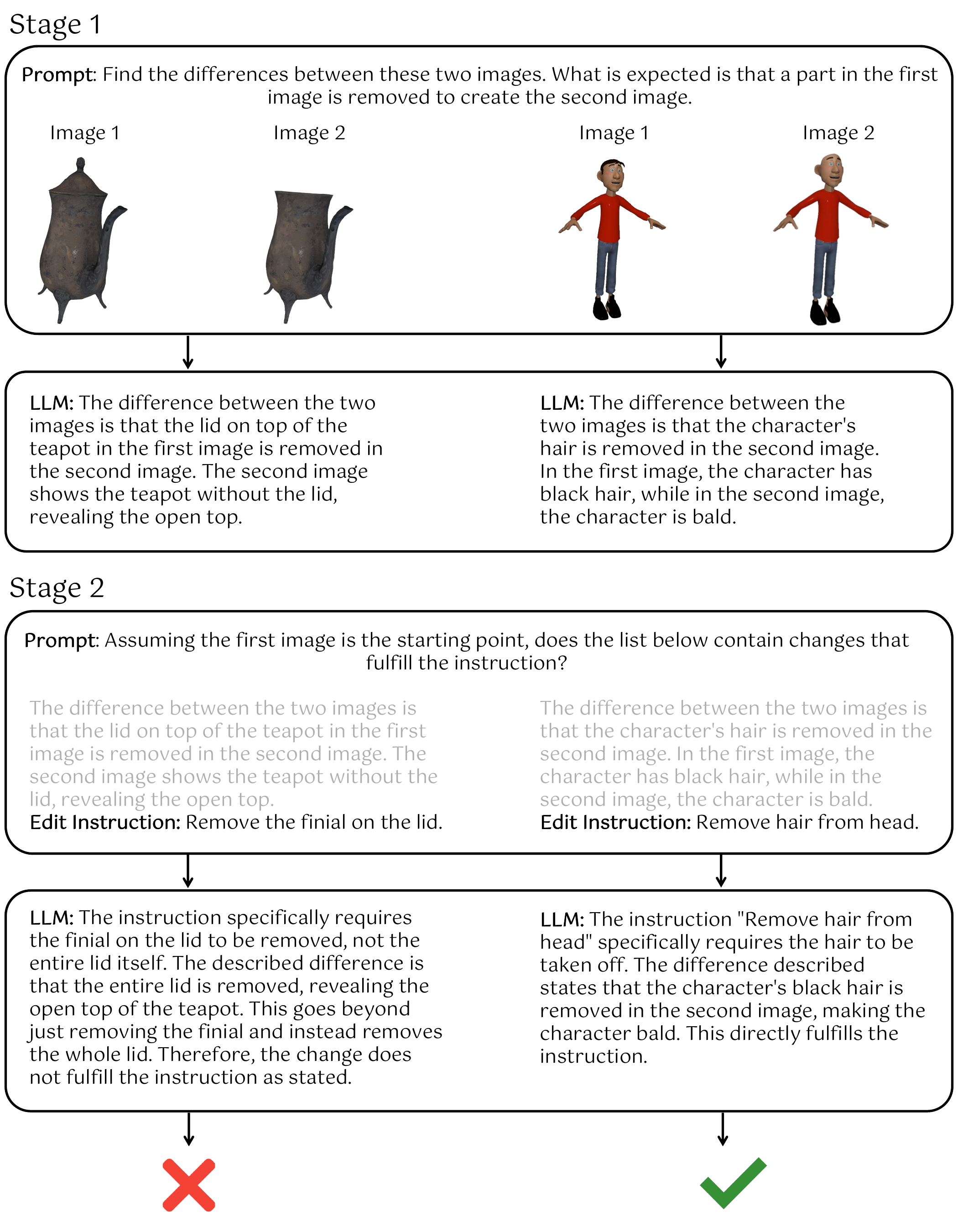}}
\caption{LLM-based filter to filter out image editing errors. One LLM, without knowing the editing instruction, is asked to list the differences between two renderings of the object from the same camera angle. Another LLM compares whether the difference matches exactly with the edit instruction. Objects that fail this check are rejected.}
\label{fig:llmfilter}
\end{center}
\end{figure*}

To address this challenge, we use two LLMs collaboratively.
As shown in \cref{fig:llmfilter}, one LLM is instructed to list the differences between two renderings of the object from the same camera angle, without knowing the editing instruction. Another LLM, without seeing the images, compares whether the difference description generated by the first LLM aligns exactly with the editing instruction. This method effectively reduces hallucination and improves filtering effectiveness, thus ensuring better data quality.

\begin{figure*}[ht]
\begin{center}
\centerline{\includegraphics[width=1.5\columnwidth]{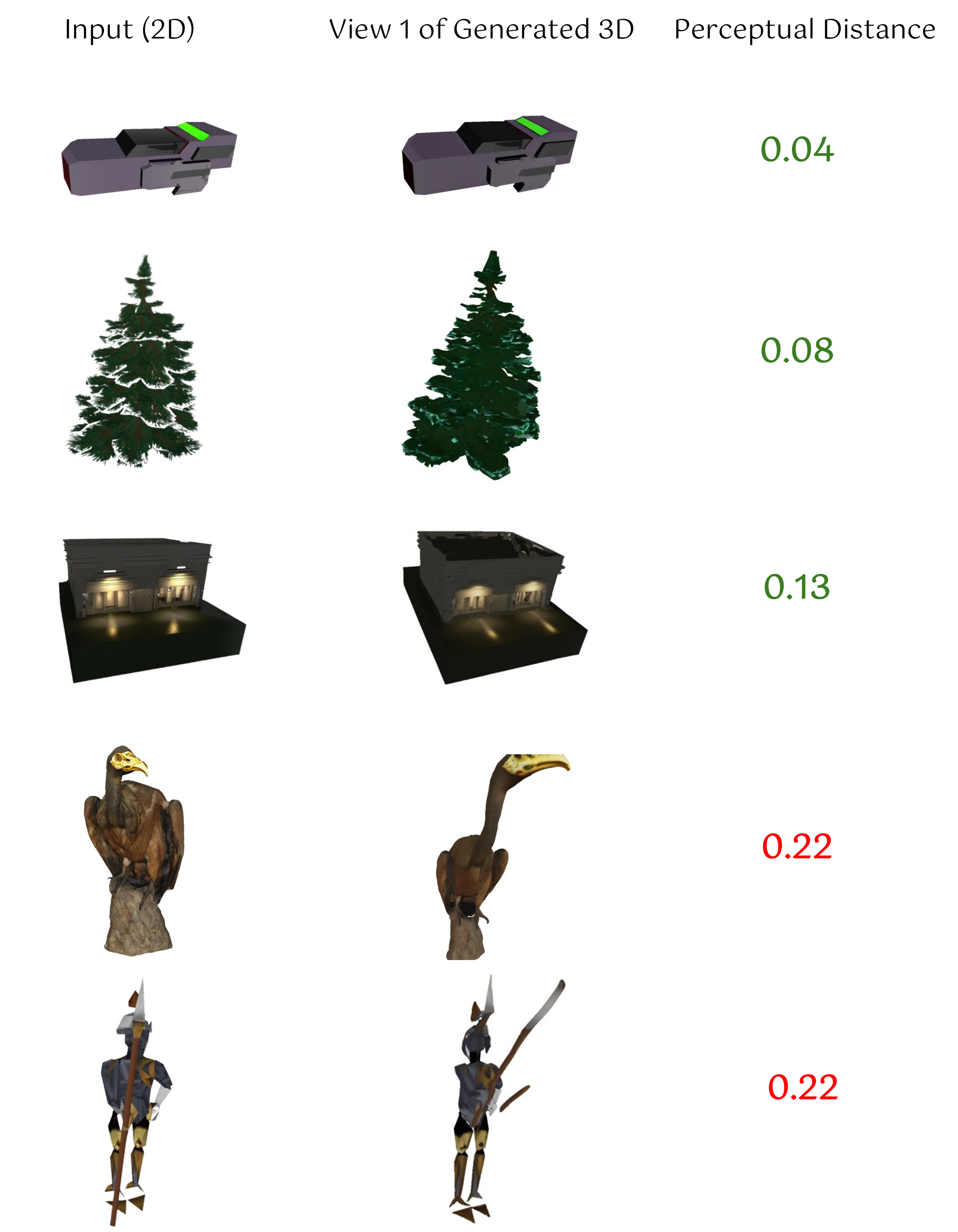}}
\caption{2D perceptual similarity (DreamSim) score to filter out reconstruction inconsistency. Consistent objects yield a low distance, and inconsistent objects yield a high distance.}
\label{fig:dreamsim}
\end{center}
\end{figure*}

\mypar{2D Perceptual Similarity for Consistency. }
We want the edited object to not only follow the text instruction correctly, but also keep consistent with the original object (modulo the edit). This means that we do not want big fluctuations in size, shape, geometry, or orientation. To achieve this, we employ a 2D perceptual similarity metric, DreamSim~\citep{fu_dreamsim_2023}. \cref{fig:dreamsim} shows some examples that pass or fail the perceptual check. We note that these metrics cannot be used as the sole determinator of editing consistency -- a ``no edit" would yield a perfect consistency distance of 0, despite being incorrect. However, given that the examples already pass through the LLM-based checking described in the previous paragraph, we know the editing has been carried out correctly, and thus perceptual distance can be used here as a proxy for consistency.

\section{Author Contributions}
ZM conceived the project and developed the model and training recipe. HC implemented and developed the data engine and benchmark.

\end{document}